%% file: main.tex
\documentclass[conference]{IEEEtran}
\IEEEoverridecommandlockouts
\usepackage{cite}
\usepackage{amsmath,amssymb,amsfonts}
\usepackage{graphicx}
\usepackage{textcomp}
\usepackage{xcolor}
\usepackage{diagbox}
\usepackage{enumerate}

\usepackage[ruled,vlined,linesnumbered]{algorithm2e}

\newtheorem{definition}{\textbf{Definition}}
\newtheorem{theorem}{\textbf{Theorem}}
\newtheorem{lemma}{\textbf{Lemma}}

\renewcommand{\IEEEQED}{\IEEEQEDopen}

\makeatletter
\newcommand{\removelatexerror}{\let\@latex@error\@gobble}
\newcommand\figcaption{\def\@captype{figure}\caption}
\newcommand\tabcaption{\def\@captype{table}\caption}
\makeatother

\def\BibTeX{{\rm B\kern-.05em{\sc i\kern-.025em b}\kern-.08em
    T\kern-.1667em\lower.7ex\hbox{E}\kern-.125emX}}
\begin{document}
\title{Rank-Regret Minimization}

\author{\IEEEauthorblockN{Xingxing Xiao$^{\hspace{.1em}\S\ddag}$, Jianzhong Li$^{\hspace{.1em}\ddag\S}$}
	\vspace{.4em}
	\IEEEauthorblockA{\textit{$^\S$Department of Computer Science and Technology, Harbin Institute of Technology, Harbin, China} \\
		\textit{$^\ddag$Faculty of Computer Science and Control Engineering, Shenzhen Institute of Advanced Technology} \\ 
		\textit{Chinese Academy of Sciences, Shenzhen, China} \\
		\vspace{4pt}
		\{xiaoxx, lijzh\}@hit.edu.cn}
}

\maketitle

\input{file/abstract}

\input{file/introduction}
\input{file/definition}

\input{file/theoretic}
\input{file/2D}
\input{file/HD}

\input{file/experiments}
\input{file/conclusion}

\section*{Acknowledgments}
This work is supported by the National Natural Science Foundation of China (NSFC) Grant NOs. 61732003, 61832003, 61972110, U1811461 and U19A2059, and the National Key R\&D Program of China Grant NO. 2019YFB2101900.

\bibliographystyle{plain}
\bibliography{sample.bib}

\end{document}

%% file: file/abstract.tex
\begin{abstract}
Multi-criteria decision-making often requires finding a small representative set from the database. A recently proposed method is the regret minimization set (RMS) query. RMS returns a size $r$ subset $S$ of dataset $D$ that minimizes the regret-ratio (the difference between the score of top-1 in $S$ and the score of top-1 in $D$, for any possible utility function). 
RMS is not shift invariant, causing inconsistency in results. Further, existing work showed that the regret-ratio is often a “made up” number and users may mistake its absolute value. Instead, users do understand the notion of rank. Thus it considered the problem of finding the minimal set $S$ with a rank-regret (the rank of top-1 tuple of $S$ in the sorted list of $D$) at most $k$, called the rank-regret representative (RRR) problem. 

Corresponding to RMS, we focus on the min-error version of RRR, called the rank-regret minimization (RRM) problem, which finds a size $r$ set to minimize the maximum rank-regret for all utility functions. Further, we generalize RRM and propose the restricted RRM (i.e., RRRM) problem to optimize the rank-regret for functions restricted in a given space. Previous studies on both RMS and RRR did not consider the restricted function space. The solution for RRRM usually has a lower regret level and can better serve the specific preferences of some users.
Note that RRM and RRRM are shift invariant. In 2D space, we design a dynamic programming algorithm 2DRRM to return the optimal solution for RRM. In HD space, we propose an algorithm HDRRM that introduces a double approximation guarantee on rank-regret. Both 2DRRM and HDRRM are applicable for RRRM. Extensive experiments on the synthetic and real datasets verify the efficiency and effectiveness of our algorithms. In particular, HDRRM always has the best output quality in experiments.
\end{abstract}

\begin{IEEEkeywords} 
	Top-k query; Skyline; Multi-criteria decision-making; Regret-ratio; Rank-regret
\end{IEEEkeywords}

%% file: file/introduction.tex
\section{Introduction}
\label{s_intro}
It is a significant problem to produce a representative tuple set from the database for multi-criteria decision-making. The problem is fundamental in many applications where the user is only interested in some or even one tuple in a potentially huge database. Consider the following example. Alice visits a large car database where each car tuple has two attributes, horse power (HP) and miles per gallon (MPG). Alice is looking for a car with high MPG and high HP. There is a trade-off between these two goals, since the increase in horse power comes at the cost of reduced fuel economy. And it may be impossible for Alice to browse every car tuple. How can the database provide a representative set to best assist Alice in decision-making?

One approach is the top-$k$ query \cite{ilyas2008survey}, which requires a predefined \textit{utility function} to model user preferences. The function assigns a utility/score to each tuple in database and the query returns the $k$ tuples with highest scores. A widely used function is the weighted linear combination of tuple attributes, i.e., $\sum w_iA_i$. 
For example, Alice gives weights $70\%$ and $30\%$ to MPG and HP respectively, and the utility function is $0.7\times$MPG$+0.3\times$HP. However, Alice may roughly know what she is looking for, and it is difficult to accurately determine the function. Instead of asking users for their preferences, Borzsony et. al. \cite{borzsony2001skyline} proposed the skyline query which returns the smallest set that contains the best tuple for any monotonic utility function. However, it has the disadvantage of returning an unbounded number of tuples. 


Recently, the regret minimization set (RMS) query was proposed by Nanongkai et al. \cite{nanongkai2010regret} to address the issues of previous methods. It needs no feedback from users and its output size is controllable. RMS aims to find a size $r$ subset $S$ of the dataset $D$ such that for any user, the top-$1$ tuple of $S$ is a good approximation of the top-$1$ of $D$. Nanongkai defined the \textit{regret-ratio} to measure the “\textit{regret}” level of a user if s/he gets the best of $S$ but not the best in $D$. Given a function $f$, let $w$ be the highest utility of tuples in $D$ and $w'$ be that of $S$. Then the regret-ratio for $f$ is $(w-w')/w$. And RMS returns a size $r$ set that minimizes the maximum regret-ratio for all possible utility functions.

Nevertheless, the RMS query and regret-ratio have some shortcomings. First, RMS is not \textit{shift invariant} (i.e., if we add a fixed constant to all values in some attribute, RMS may change the result), which causes inconsistency in outputs. There are attributes (e.g., temperature) that shift values when converting between scales (e.g., °C, °F, and K). Similarly, the altitudes and spatial coordinates are shifted as the reference point changes. After shifting, the dataset is essentially unchanged and query results should be the same. Paradoxically, RMS may have different outputs.
Further, RMS assumes a larger value is preferable in each attribute and all values are non-negative. To make the assumption satisfied, the data has to be shifted. For smaller-preferred attributes (e.g., price), RMS negates values in them and yields some negative ones. For negative values (e.g., temperature), it shifts tuples to eliminate them. However, the shift degree seriously affects the results of RMS and introduces inconsistency.

Second, as shown in \cite{asudeh2019rrr}, the utility/regret-ratio is usually an artificial number with no direct significance, and users may not understand its absolute value. For example, wine ratings are on a 100-point scale. Wines rated below 80, which have regret-ratios about $0.2$ and seem attractive to novice drinkers, are almost never officially sold in stores. On the shopping website Taobao, the logistics score of Adidas flagship store is 4.8 points (out of 5), which is the median but guarantees a regret-ratio less than $0.04$. The average logistics level on Taobao is far worse than that on its competitor JD.com. It is also shown that a small range in regret-ratio can include a large fraction of dataset.

Relative to the regret-ratio, users do understand the notion of rank. Consequently, Asudeh et al. \cite{asudeh2019rrr} measured the user regret by the rank. Given a function $f$, they defined the \textit{rank-regret} of a set $S$ to be the rank of the best tuple of $S$ in the sorted list of dataset $D$ w.r.t. $f$. Obviously, if the rank-regret of $S$ for any possible utility function is at most $k$, then it contains a top-$k$ tuple for each user. For example, a subset of wines with a rank-regret of 6 should contain one of the top-6 wines in the mind of any user, which is close to the top-1.  


Specifying the output size is critical due to some considerations such as the website display size, upload speed to the cloud and communication bandwidth between hosts. In this paper, we investigate the rank-regret minimization (RRM) problem that finds a size $r$ set to minimize the maximum rank-regret for all utility functions. Further, we generalize RRM and propose the restricted rank-regret minimization (RRRM) problem with an additional input, the restricted function space $\mathbb{U}$. RRRM no longer pays attention to all possible functions and aims to minimize the rank-regret for any function in $\mathbb{U}$. $\mathbb{U}$ may be the candidate space of function used by a user. Specified directly by the user or mined by a learning algorithm\cite{joachims2002optimizing, wang2011latent, jamieson2011active, qian2015learning}, the obtained function is inherently inaccurate. However, as shown in \cite{ciaccia2017reconciling, mouratidis2018exact, mouratidis2021marrying, liu2021eclipse}, it can be used as a rough guide and expanded into a candidate space $\mathbb{U}$. In addition, $\mathbb{U}$ can be designated by experts and the functions not in $\mathbb{U}$ are impossible for users to use. Under the same settings, the solution of RRRM usually has a lower rank-regret than RRM, owing to fewer functions in $\mathbb{U}$. It can better serve the specific preferences of some users. Contrary to RMS, both RRM and RRRM are shift invariant.

Asudeh et al. \cite{asudeh2019rrr} focused on the dual formulation of RRM, the rank-regret representative (RRR) problem. Given a threshold $k$, it finds the minimum set with a rank-regret at most $k$. In 2D space, they proposed an $O(n^2\log n)$ time and $O(n^2)$ space approximation algorithm 2DRRR that returns a set with size and rank-regret at most $r_k$ and $2k$ respectively, where $r_k$ is the minimal size of sets with rank-regrets at most $k$. 
In high-dimensional (HD) space (i.e., dimension $d>2$), based on the combinatorial geometry notion of $k$-set \cite{edelsbrunner2012algorithms}, they presented an $O(|W|kn\mathrm{LP}(d, n))$ time and $O(|W|k)$ space algorithm MDRRR with a logarithmic approximation-ratio on size and a rank-regret of $k$, where $\mathrm{LP}(d, n)$ is the time for solving a linear programming with $d$ variables and $n$ constrains, and $|W|$ is the number of $k$-sets (its best-known lower-bound is $n^{d-1}e^{\Omega(\sqrt{\log n})}$ \cite{toth2000point}). As shown in \cite{asudeh2019rrr}, MDRRR is quite impractical and does not scale beyond a few hundred tuples. Therefore, they showed a randomized version, MDRRR$_r$, which reduces the time complexity to $O(|W|(nd+k\log k))$ but no longer has a guaranteed rank-regret. Finally, they gave a heuristic algorithm MDRC based on space partitioning. Asudeh et al. \cite{asudeh2019rrr} didn't consider the restricted function space.


In this paper, we present several theoretical results as well as practical advances for RRM and RRRM, in both 2D and HD cases. In 2D space, we design a dynamic programming algorithm 2DRRM to return the optimal solution for RRM, indicating that RRM is in P for $d=2$. It can be transformed into an exact algorithm for RRR and is a huge improvement over 2DRRR\cite{asudeh2019rrr}. In HD space, we discretize the continuous function space into a bounded size set of utility functions and prove from two different perspectives that if a tuple set has a small rank-regret for functions in the discretized set, then it is approximately the same for the full space. We then convert the problem into linear number of set-cover \cite{bronnimann1995almost} instances. This leads to the design of the algorithm HDRRM, which has a double approximation guarantee on rank-regret. The output of HDRRM always has the lowest rank-regret in experiments. Both 2DRRM and HDRRM can be generalized to RRRM. 

The main contributions of this paper are listed below.
\begin{enumerate}[$\bullet$]
	\item We generalize RRM and propose the RRRM problem, which aims to minimize the regret level for any utility function in a restricted space. Under the same settings, the solution of RRRM usually has a better quality. 
	\item We prove that RRM and RRRM are shift invariant. Second, we provide a lower-bound $\Omega(n/r)$ on rank-regret, indicating that there is no algorithm with a rank-regret upper-bound independent of the data size $n$. In addition, we show that the restricted skyline\cite{ciaccia2017reconciling} is a set of candidate tuples for RRRM. Relatively, skyline tuples\cite{borzsony2001skyline} are candidates for RRM.
	\item In 2D space, we design an $O(n^2\log n)$ time and $O(n^2)$ space algorithm 2DRRM to return the optimal solution for RRM. It can be used to find the optimal solution for RRR and be applied for RRRM through some modifications.
	\item In HD space, we propose an $O(n\log^2 n)$ time and $O(n\log n)$ space algorithm HDRRM for RRM that returns a size $r$ set and approximates the minimal rank-regret for any $O(\frac{r}{\ln\ln n})$ size set. It is the only HD algorithm that has a rank-regret guarantee and is suitable for RRRM. HDRRM always has the best output quality in experiments.
	\item Extensive experiments on the synthetic and real datasets verify the efficiency and effectiveness of our algorithms.
\end{enumerate}

%% file: file/definition.tex
\section{PROBLEM DEFINITION}
\label{s_def}

Let $D$ be a dataset containing $n$ tuples with $d$ numeric attributes and $A_i$ be the $i$-th attribute. Given a tuple $t$, its value on $A_i$ is denoted by $t[i]$. The L2-norm of $t$ is abbreviated as $\Vert t\Vert=(\sum_{i=1}^d t[i]^2)^\frac{1}{2}$. Assume that on each attribute, a larger value is preferred and the range is normalized to $[0,1]$. For each $A_i$, there is a tuple $t\in D$ with $t[i] = 1$, named as the $i$-th dimensional \textit{boundary tuple}. 
Define the \textit{basis} \cite{agarwal2017efficient} of $D$, denoted by $B$, to be the set of all boundary tuples. Assume that the dimensionality $d$ is a fixed constant, which is reasonable in many scenarios and appears in many related works (e.g., \cite{asudeh2017efficient, asudeh2019rrr, xie2018efficient, xie2020being, xie2020experimental}). Given an integer $k\ge 1$, let $[k]$ denote the integer set $\{1, 2, \dots, k\}$. 

Suppose the user preference is modeled by an unknown utility function, which assigns a non-negative utility/score $f(t)$ to each tuple $t\in D$. To avoid several complicated but uninteresting "boundary cases", assume that no two tuples have the same utility in $D$. Following \cite{wang2021minimum, xie2020being, asudeh2019rrr, asudeh2017efficient, agarwal2017efficient, wang2021fully}, focus on the popular-in-practice linear utility functions, shown to effectively model the way users evaluate trade-offs in real-life multi-objective decision-making \cite{qian2015learning}. A utility function $f$ is linear, if 
\vspace{-2ex} $$f(t) = w(u,t) = \sum_{i=1}^{d}u[i] t[i]\vspace{-1ex}$$ 
where $u=(u[1], \cdots , u[d])$ is a $d$-dimensional non-negative real vector and $u[i]$ measures the importance of attribute $A_i$. In the following, refer $f$ by its utility vector $u$ and use them interchangeably. A tuple $t$ outranks a tuple $t'$ based on $u$, if $w(u,t)>w(u,t')$. A user wants a tuple which maximizes the utility w.r.t. his/her utility function. Given a utility vector $u$ and an integer $k\ge 1$, let $w_k(u, D)$ be the $k$-th highest utility of tuples in $D$. For brevity, $w(u, D) = w_1(u, D)$. The tuple $t\in D$ with $w(u,t) = w(u,D)$ is called the \textit{highest utility tuple} of $D$ for $u$. Set 
\vspace{-1ex} $$\Phi_k(u, D) = \{t\in D\ |\ w(u,t)\ge w_k(u,D)\} \vspace{-1ex}$$ 
to be the set of top-$k$ tuples w.r.t. $u$. For each $t \in D$, $\nabla_u(t)$ denotes the rank of $t$ in the sorted list of $D$ in descending order based on $u$. There are exactly $\nabla_u(t)-1$ tuples in $D$ that outrank $t$ according to $u$. Through $\nabla_u(t)$, $\Phi_k(u, D)$ is also defined as the set $\{t\in D\ |\ \nabla_u(t)\le k\}$. Next, define the rank-regret of a tuple set for a given utility vector. 

\begin{definition}
	Given a set $S\subseteq D$ and a utility vector $u$, the rank-regret of $S$ for $u$, denoted by $\nabla_u(S)$, is the minimum rank of tuples in $S$ based on $u$, i.e., 
	\vspace{-1ex} $$ \nabla_u(S)=\min_{t\in S}\nabla_u(t).$$
\end{definition}
Intuitively, when the best tuple of $S$ approaches that of $D$, the rank-regret of $S$ becomes smaller, indicating that the user feels less regretful with $S$. 

In reality, it is difficult to accurately determine the utility vector. Specified directly by the user or mined by a learning algorithm, the obtained vector is inherently inaccurate. In the former case, it is impossible for users to reasonably quantify the relative importance of various attributes with absolute precision. In the latter, the mined vector is a rough estimate but not an exact representation of user preferences. A natural way is to expand a single vector into a vector space. Accordingly, define the rank-regret for a given set of utility vectors.

\begin{definition}
	Given a set $S\subseteq D$ and a set $\mathbb{U}$ of utility vectors, the rank-regret of $S$ for $\mathbb{U}$, denoted by $\nabla_\mathbb{U}(S)$, is the maximum rank-regret of $S$ for all vectors in $\mathbb{U}$, i.e., 
	\vspace{-1ex} $$ \nabla_\mathbb{U}(S) = \max_{u\in \mathbb{U}}\nabla_u(S). $$
\end{definition}
Intuitively, $\nabla_\mathbb{U}(S)$ measures the rank of best tuple of $S$ w.r.t $\mathbb{U}$ in the worst case. A set $S\subseteq D$ has $\nabla_\mathbb{U}(S)\le k$, if $\forall u\in\mathbb{U}$, $w(u,S)\ge w_k(u,D)$. Notice that $\nabla_\mathbb{U}(S)$ is a monotonic decreasing function. Given two sets $R$ and $S$, if $R\subseteq S\subseteq D$, then $\nabla_\mathbb{U}(R)\ge \nabla_\mathbb{U}(S)$. 

With no knowledge about user preferences, our goal is to find a given size set with a small rank-regret for any linear utility function. Intuitively, the class of linear functions corresponds to the $d$-dimensional non-negative orthant 
\vspace{-1ex} $$\mathbb{L} = \{(u[1], \dots, u[d])\ |\ \forall i\in [d], u[i] \ge 0\}. \vspace{-1ex} $$ 
Consequently, we define the rank-regret minimization (RRM) problem as follows.

\begin{definition}[RRM Problem]
	Given a dataset $D$ and an integer $r\ge1$, compute a size $r$ subset of $D$ that minimizes the rank-regret for $\mathbb{L}$, i.e., return a set 
	\vspace{-1ex} $$ S^* = \mathop{\arg\min}\limits_{S\subseteq D: |S|\le r}\nabla_\mathbb{L}(S).$$ 
\end{definition}

Through some prior knowledge about user preferences \cite{ciaccia2017reconciling, mouratidis2018exact, mouratidis2021marrying, liu2021eclipse}, assume that the utility vector lies in a restricted space and propose the restricted RRM problem.

\begin{definition}[RRRM Problem]
	Given a dataset $D$, a set of utility vectors $\mathbb{U}\subseteq\mathbb{L}$ and an integer $r\ge1$, compute a size $r$ subset of $D$ that minimizes the rank-regret for $\mathbb{U}$, i.e., return a set 
	\vspace{-1ex} $$ S^* = \mathop{\arg\min}\limits_{S\subseteq D: |S|\le r}\nabla_\mathbb{U}(S).$$ 
\end{definition}
When $\mathbb{U} = \mathbb{L}$, RRRM degenerates into RRM. The most relevant researches on the restricted space are \cite{ciaccia2017reconciling, mouratidis2018exact, mouratidis2021marrying, liu2021eclipse}. $\mathbb{U}$ was assumed as a convex polytope\cite{ciaccia2017reconciling, mouratidis2018exact}, a hyper-sphere\cite{mouratidis2021marrying} or an axis-parallel hyper-rectangle\cite{liu2021eclipse}. They did not consider minimizing rank-regret and controlling the output size. In this paper, we assume that $\mathbb{U}$ can be any convex space, which is more generalized than the previous researches.


\begin{figure*}[tb]
	\centering
	\begin{minipage}[tb]{0.37\linewidth}
		\tabcaption{A 2D dataset} 
		\centering
		\label{tab_example}
		\footnotesize
		\begin{tabular}{c|c|c|c|c}
			& $A_1$ & $A_2$ & Rank-Ratio & Regret-Ratio \\ \hline
			$t_1$ & 0 & 1 & 7 & 100\% \\ \hline
			$t_2$ & 0.4 & 0.95 & 4 & 60\% \\ \hline
			$t_3$ & 0.57 & 0.75 & 3 & 43\% \\ \hline
			$t_4$ & 0.79 & 0.6 & 4 & 40\% \\ \hline
			$t_5$ & 0.2 & 0.5 & 6 & 80\% \\ \hline
			$t_6$ & 0.35 & 0.3 & 6 & 70\% \\ \hline
			$t_7$ & 1 & 0 & 7 & 100\%
		\end{tabular}
	\end{minipage}
	\begin{minipage}[tb]{0.2\linewidth}
		\centering
		\includegraphics[width=1\linewidth]{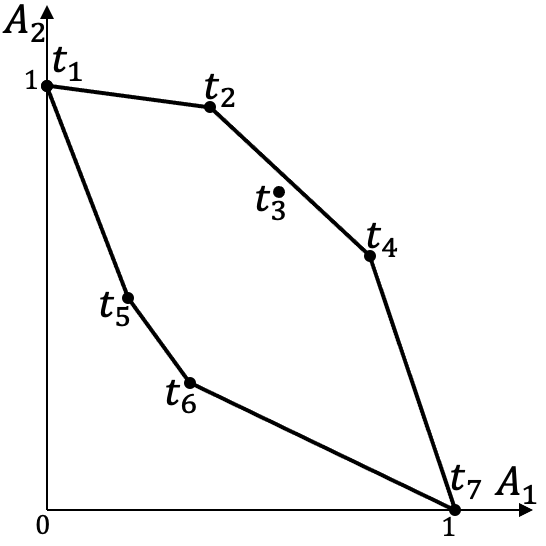} 
		\caption{Tuples of Table \ref{tab_example} in 2D space}
		\label{fig-example}
	\end{minipage}
	\begin{minipage}[tb]{0.2\linewidth}
		\centering
		\includegraphics[width=1\linewidth]{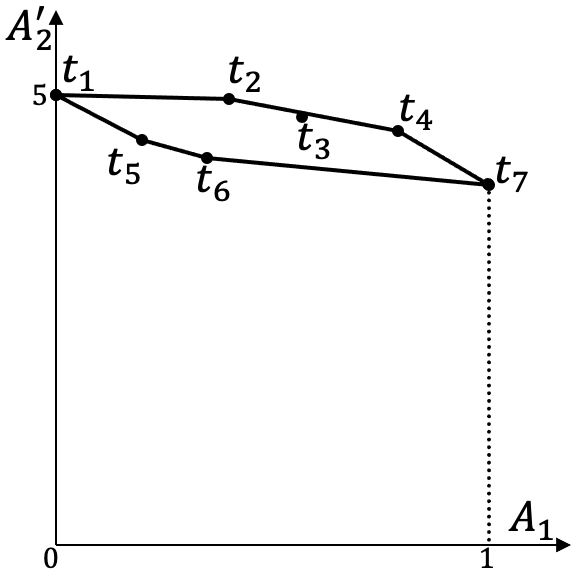} 
		\caption{Shifted Tuples}
		\label{fig-shift}
	\end{minipage}
	\begin{minipage}[tb]{0.2\linewidth}
		\centering
		\includegraphics[width=1\linewidth]{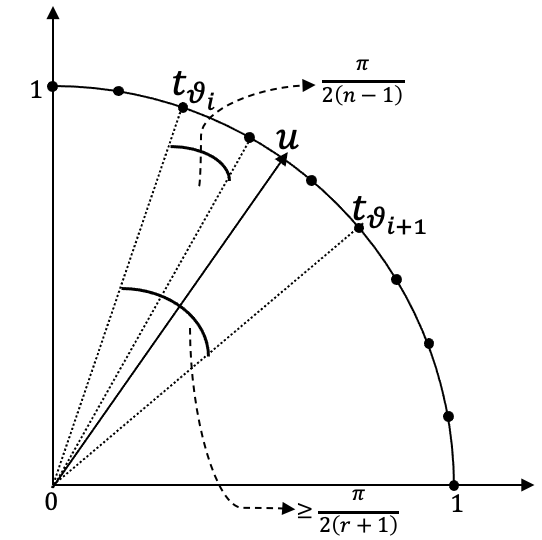} 
		\caption{Proof Idea of Theorem \ref{t_low_bound}}
		\label{fig-lower_bound}
	\end{minipage}
	\vspace{-3ex} 
\end{figure*}

The solution of RRM is close to that of RMS\cite{nanongkai2010regret} when tuples are uniformly distributed by utility. Table \ref{tab_example} shows a dataset with $7$ tuples over two attributes and Figure \ref{fig-example} shows them as points in 2D space. When $r=1$, the solutions for RRM and RMS are $\{t_3\}$ and $\{t_4\}$ respectively. In terms of rank-regrets and regret-ratios, the two are close. However, tuples may be clustered in a small utility range. RMS pursues approximations on the absolute utility (i.e., regret-ratio), possibly resulting in a significant increase on rank-regret. In Table \ref{tab_example}, add $4$ to each value on attribute $A_2$ and obtain the shifted tuples shown in Figure \ref{fig-shift}. Since the updated values are in the top range, RMS directly ignores $A_2$ and seeks the largest value on $A_1$, i.e., the result of RMS becomes $\{t_7\}$, indicating that RMS is not shift invariant. But $t_7$ has the worst rank on $A_2$. And the solution of RRM is still $\{t_3\}$. In short, minimizing regret-ratio does not typically minimize rank-regret.


Asudeh et al. \cite{asudeh2019rrr} focused on the dual version of RRM, called RRR, i.e., find the minimum set that fulfill a given rank-regret threshold. Depending on the search needs of users, different versions can be applied. A solver for RRM can be easily adopted for RRR by a binary search with an additional $\log n$ factor in the running time. The introduction shows several situations where it is critical to specify the output size. The output size also represents the effort required by the user to make a decision. RRR cannot directly control the output size and may return a huge set. Users may be unable or unwilling to browse thousands of tuples. A large representative set provides little help for decision-making. Therefore, we usually would like to return only a limited number $r$ of tuples.

For a huge dataset, due to the range $[1,n]$, RMS/RRMS may have a solution with a rank-regret that seems large. But in fact the regret level of the user is not high. For datasets of different sizes, the same rank-regret has different practical meanings. Similar to regret-ratio, we can normalize the range to $[0,1]$ (i.e., divide rank-regrets by $n$) and equivalently use percentages to represent user regret. Expressing rank in the form of percentage is much common. For example, highly cited papers are usually defined as those that rank in the top $1\%$ by citations in the field and publication year.

\textbf{Complexity Analysis:} Asudeh et al. \cite{asudeh2019rrr} proved that RRM is NP-Complete for $d\ge 3$. Since RRM is the sub-problem of RRRM, RRRM is NP-Hard for $d\ge 3$. In this paper, we propose a polynomial time exact algorithm for RRM when $d=2$, which is applicable for RRRM. It supplements the previous analysis and shows that in 2D space, RRM and even RRRM are in P.

%% file: file/theoretic.tex
\section{Theoretical Property}
\label{s_theoretical} 
In this section, we introduce a few theoretical properties about the RRM and RRRM problems. First, we claim that both problems are shift invariant. Second, by proving a lower-bound on the rank-regret, we show that there is no algorithm for RRM with an upper-bound guarantee independent of the data size. Last, we clarify that the restricted skyline \cite{ciaccia2017reconciling} is the set of candidate tuples for RRRM. Correspondingly, the \textit{skyline} \cite{xiao2020sampling} is the candidate set for RRM.

\subsection{Shift Invariance}
Given two datasets $D = \{t_1, ..., t_n\}$ and $D' = \{t'_1, ..., t'_n\}$, $D'$ is a shifting of $D$, if there are $d$ non-negative reals $\lambda_1$, $...$, $\lambda_d$ such that $\forall i\in[n]$ and $\forall j\in[d]$, $t'_i[j] = \lambda_j+t_i[j]$. Shift invariant means that for any dataset $D$, shifting does not change the optimal solution of problem. As shown in Section \ref{s_def}, RMS is not shift invariant. We now claim that both RRM and RRRM satisfy the property.

\begin{theorem}
	RRM and RRRM are shift invariant. 
\end{theorem}
\begin{IEEEproof} Consider a dataset $D = \{t_1, ..., t_n\}$. Suppose that $D' = \{t'_1, ..., t'_n\}$ is a shifting of $D$, i.e., there are positive reals $\lambda_1, \lambda_2, ..., \lambda_d$ such that $\forall i\in[n]$ and $\forall j\in[d]$, $t'_i[j] = \lambda_j+t_i[j]$. For any $u\in\mathbb{U}$, $w(u,t'_i) = \sum_{j=1}^{d}u[j]t'_i[j] = \sum_{j=1}^{d}u[j]\lambda_j+\sum_{j=1}^{d}u[j]t_i[j] = \sum_{j=1}^{d}u[j]\lambda_j+w(u,t_i)$. Thus the rank of $t'_i$ in $D'$ w.r.t. $u$ is the same as that of $t_i$ in $D$. And the rank-regret of any set w.r.t. $u$ remains the same after shifting. So does the rank-regret w.r.t. $\mathbb{U}$. Thus RRM (let $\mathbb{U}$ be $\mathbb{L}$) and RRRM are shift invariant. \end{IEEEproof}

For lack of space, the missing proofs in this paper can be found in \cite{xiao2021rankregret}.

\subsection{Lower-Bound}
\label{s_lower_bound}

For RMS, Xie et al. \cite{xie2018efficient} showed that there is a dataset over $d$ attributes such that the minimum regret-ratio for any size $r$ set is $\Omega(r^{-2/(d-1)})$. Correspondingly, we show a lower-bound on the rank-regret.
\begin{theorem}
	\label{t_low_bound}
	Given integers $r\ge1$ and $d\ge2$, there is a dataset $D$ of $n$ $d$-dimensional tuples such that for any size $r$ set $S\subseteq D$, $\nabla_\mathbb{L}(S)$ is $\Omega(n/r)$. 
\end{theorem}
\begin{IEEEproof}
	First, assume $d=2$. Our construction is inspired by Nanongkai et al. \cite{nanongkai2010regret}. We define the angle of a tuple/point $t$ to be $\arctan\frac{t[2]}{t[1]}$. Let $D$ be the size $n$ set $$ \{t_\theta = (\cos\theta, \sin\theta)\ |\ \theta\in\{0, \frac{\pi}{2(n-1)}, \cdots, \frac{(n-1)\pi}{2(n-1)}\}\}. $$ All tuples (black dots in Figure \ref{fig-lower_bound}) lie on a quarter-arc in $\mathbb{R}^2_+$ with radius $1$ centered at the origin. Then the angle of $t_\theta$ is $\theta$. Each tuple in $D$ is the highest utility tuple for some vectors in $\mathbb{L}$. Consider a size $r$ set $S\subseteq D$. Let the angles of tuples in $S$ are $\vartheta_1 < \vartheta_2 <\cdots < \vartheta_r$. Let $\vartheta_0=0$ and $\vartheta_{r+1}=\pi/2$, i.e., $t_{\vartheta_0}=(1,0)$ and $t_{\vartheta_{r+1}}=(0,1)$. Let $p$-$O$-$q$ be the angle subtended between lines obtained by joining the origin $O = (0, 0)$ with two points $p$ and $q$. Consider angles $t_{\vartheta_i}$-$O$-$t_{\vartheta_{i+1}}$ for all $i\in\{0,\cdots, r\}$. Note that the sum of these $r+1$ angles is exactly $\pi/2$. Therefore, at least one angle is no less than $\frac{\pi}{2(r+1)}$. Let the angle be $t_{\vartheta_i}$-$O$-$t_{\vartheta_{i+1}}$. The difference between angles of two adjacent tuples in $D$ is $\frac{\pi}{2(n-1)}$. Thus there are $\Omega(n/r)$ tuples in $D$ whose angles are larger than $\vartheta_i$ but less than $\vartheta_{i+1}$. If $i\in [r-1]$, then set $\alpha=\frac{\vartheta_i+\vartheta_{i+1}}{2}$ and consider the utility vector $u=(\cos\alpha, \sin\alpha)$. Note that the utility of each tuple $t\in D$ based on $u$ is the $L_2$-distance between the origin $O$ and the projection of $t$ onto line $O$-$(\cos\alpha, \sin\alpha)$. And the tuples in $D$ whose angles lie between $\vartheta_i$ and $\vartheta_{i+1}$ have utilities larger than those of $t_{\vartheta_i}$ and $t_{\vartheta_{i+1}}$, i.e., outrank $t_{\vartheta_i}$ and $t_{\vartheta_{i+1}}$ based on the utility vector $u$. See Figure \ref{fig-lower_bound}. In addition, $t_{\vartheta_i}$ and $t_{\vartheta_{i+1}}$ are the top-2 tuples of $S$ for $u$. Thence the rank-regret of $S$ for $u$ is $\Omega(n/r)$. If $i = 0$, then set $\alpha=0$. If $i = r$, set $\alpha=\pi/2$. With the same analysis, $\nabla_u(S)$ is also in $\Omega(n/r)$, and so is $\nabla_\mathbb{L}(S)$. 
	
	If $d>2$, for each tuple in $D$, keep the values of first two attributes, and set those of the subsequent attributes to $1$. Then we are able to get the same result.
\end{IEEEproof}

The difference between the bounds for RMS and RRM is not just the difference in scale ($[0,1]$ for regret-ratio and $[1,n]$ for rank-regret). When $d=2$, multiplying $\Omega(r^{-2/(d-1)})$ by $n$ results in a looser bound $\Omega(n/r^2)$. For RMS, there are several algorithms \cite{nanongkai2010regret, cao2017k, agarwal2017efficient, xie2018efficient} with an upper-bound on the regret-ratio, independent of $n$. It is obvious that there is no such algorithm for RRM and RRRM. However, the bound in Theorem \ref{t_low_bound} is for the adversarial cases far from practice. Algorithms proposed in this paper usually have small rank-regrets in experiments. Further, the rank-regret can be represented by a percentage as shown before.

\subsection{Candidate Tuples}
Based on the restricted space $\mathbb{U}$, Ciaccia et al. \cite{ciaccia2017reconciling} proposed a new kind of dominance, through which they introduced the restricted version of skyline. 
\begin{definition}[$\mathbb{U}$-Dominance and $\mathbb{U}$-Skyline]
	Given $t, t'\in D$ and $\mathbb{U}\subseteq\mathbb{L}$, $t$ $\mathbb{U}$-dominates $t'$, written $t\prec_\mathbb{U} t'$, if $\forall u\in\mathbb{U}$, $w(u,t)\ge w(u,t')$ and $\exists v\in\mathbb{U}$, $w(v,t)>w(v,t')$. The $\mathbb{U}$-skyline of $D$ is 
	$\mathrm{Sky}_\mathbb{U}(D) = \{ t\in D\ |\not\exists t'\in D, t'\prec_\mathbb{U} t \}.$
\end{definition}

When $\mathbb{U} = \mathbb{L}$, $\mathrm{Sky}_\mathbb{U}(D)$ is the skyline\cite{xiao2020sampling} of $D$, abbreviated as $\mathrm{Sky}(D)$. And for any $\mathbb{U}\subseteq\mathbb{L}$, $\mathrm{Sky}_\mathbb{U}(D) \subseteq \mathrm{Sky}(D)$. We show that $\mathrm{Sky}_\mathbb{U}(D)$ is the set of candidate tuples for RRRM. 

\begin{theorem}
	\label{t_skyline}
	If there is a size $r$ set $S\subseteq D$ with $\nabla_\mathbb{U}(S) = k$, there is a set $R\subseteq\mathrm{Sky}_\mathbb{U}(D)$ with $|R| \le r$ and $\nabla_\mathbb{U}(R) \le k$.
\end{theorem}
\begin{IEEEproof}
	For each tuple $t\in S$, if $t\notin\mathrm{Sky}_\mathbb{U}(D)$, there must be a tuple $t'\in\mathrm{Sky}_\mathbb{U}(D)$ such that $t'$ $\mathbb{U}$-dominates $t$. For each $u\in\mathbb{U}$, $t'$ outranks $t$. Then replace all such $t$ in $S$ with $t’$, and obtain the set $R$. By definition, for each $u\in\mathbb{U}$, $\nabla_u(R)\le\nabla_u(S)$. Thus we have $\nabla_\mathbb{U}(R) \le k$. Further, $R$ is no larger than $S$. 
\end{IEEEproof}

It is sufficient to focus on the subsets of $\mathrm{Sky}_\mathbb{U}(D)$ to find the solution of RRRM. Correspondingly, $\mathrm{Sky}(D)$ is the set of candidate tuples for RRM. 

\begin{figure*}[tb]
	\begin{minipage}[tb]{0.2\linewidth}
		\centering
		\includegraphics[width=1\linewidth]{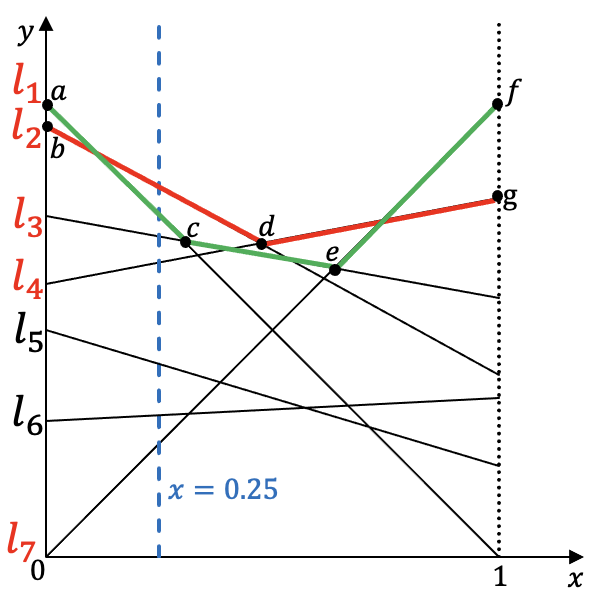} 
		\caption{Dual Representation for Tuples in Figure \ref{fig-example}}
		\label{fig-dual}
	\end{minipage}
	\begin{minipage}[tb]{0.2\linewidth}
		\centering
		\includegraphics[width=1\linewidth]{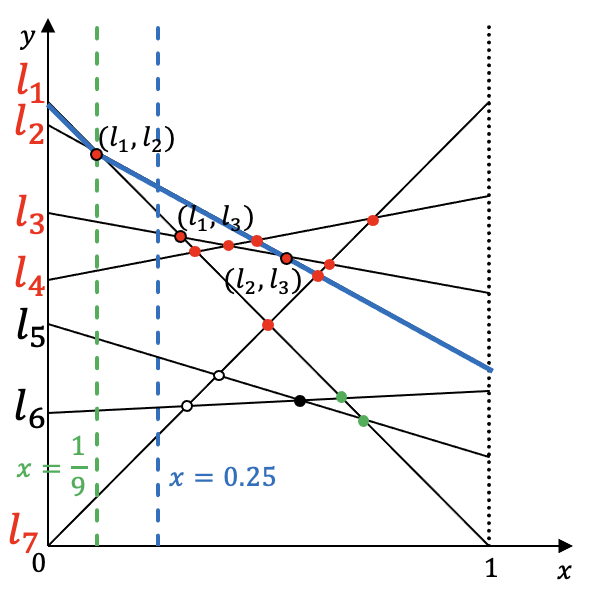} 
		\caption{Intersections between Lines}
		\label{fig-intersect}
	\end{minipage}
	\begin{minipage}[tb]{0.2\linewidth}
		\centering
		\includegraphics[width=1\linewidth]{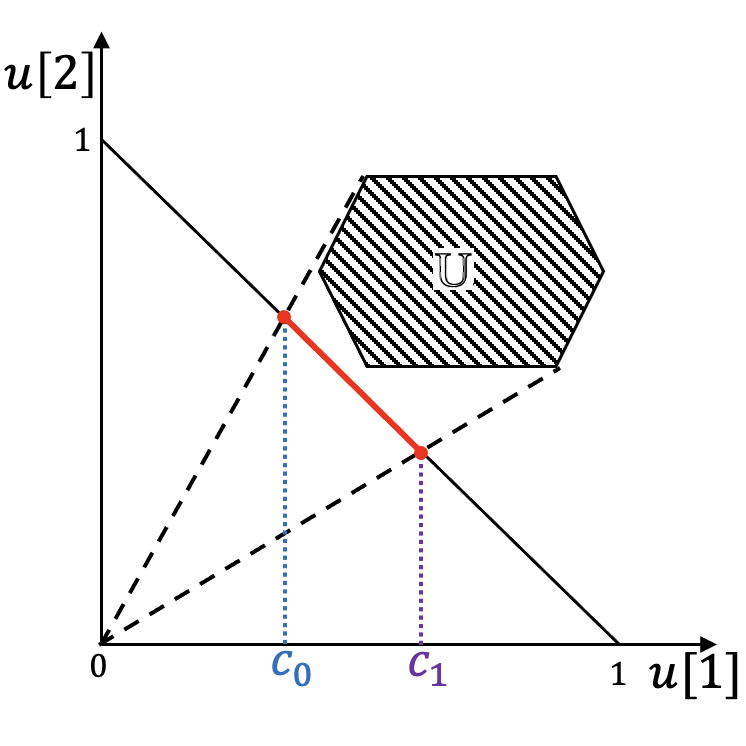} 
		\caption{Render the Scene.}
		\label{fig-render}
	\end{minipage}
	\begin{minipage}[tb]{0.37\linewidth}
		\tabcaption{updates of $M$} 
		\centering
		\footnotesize
		\label{tab_m}
		\begin{tabular}{c|c|c|c|c}
			& Initial & $(l_1,l_2)$ & $(l_1,l_3)$ & $(l_2,l_3)$\\ \hline
			$M[1,1]$ & $\{l_1\}$,1 & -,2 & -,3 & -,- \\ \hline
			$M[1,2]$ & $\{l_1\}$,1 & -,2 & -,3 & -,- \\ \hline
			$M[2,1]$ & $\{l_2\}$,2 & -,- & -,- & -,- \\ \hline
			$M[2,2]$ & $\{l_2\}$,2 & $\{l_1,l_2\}$,1 & -,- & -,2 \\ \hline
			$M[3,1]$ & $\{l_3\}$,3 & -,- & -,- & -,- \\ \hline
			$M[3,2]$ & $\{l_3\}$,3 & -,- & $\{l_1,l_3\}$,2 & -,-
		\end{tabular}
	\end{minipage}
\vspace{-3ex} 
\end{figure*}

%% file: file/2D.tex
\section{Algorithm in 2D}
\label{s_2d}
In this section, we design a dynamic programming algorithm to return the optimal solution for the RRM problem in 2D space. And we generalize it to the RRRM problem.

\subsection{Algorithm Preparation}
\label{s_2d_pre}
Inspired by Mouratidis et al. \cite{mouratidis2018exact}, in this section, assume that each utility vector $u\in\mathbb{L}$ is normalized such that $u[1] + u[2] = 1$, i.e., $u = (u[1], 1-u[1])$ for some $u[1]\in [0, 1]$. Plotting the utilities $w(u,t)$ of tuples $t=(t[1],t[2])\in D$ as functions of $u[1]$, they are each mapped into a line $l$: $ y = t[1]\cdot x+t[2]\cdot (1-x).$ 
Let $L(D)$, abbreviated as $L$, be the list of corresponding lines of tuples in $D$. Figure \ref{fig-dual} demonstrates the dual representation $L =\{l_1, l_2, \dots, l_7\}$ for dataset  $D=\{t_1, t_2, \dots, t_7\}$ in Table \ref{tab_example}. In dual space, a utility vector $u = (c, 1-c)$ ($c\in[0,1]$) corresponds to the line $x=c$. And tuple $t$ ranks higher than tuple $t'$ based on $u$, if the line of $t$ is above that of $t'$ for $x=c$. In Figure $\ref{fig-dual}$, $u=(0.25,0.75)$ is represented by the blue dashed line $x=0.25$, on which $l_2$ is above $l_1$, indicating $t_2$ outranks $t_1$. Given a line $l\in L$, the \textit{rank} of $l$ for $x=c$, denoted by $\nabla_{c}(l)$, is defined as one plus the number of lines in $L$ above $l$ for $x=c$, equal to the rank of corresponding tuple $t$ w.r.t. $u=(c,1-c)$. In Figure $\ref{fig-dual}$, the number of lines above $l_1$ for $x=0.25$ is $1$, thus $\nabla_{0.25}(l_1) = \nabla_{(0.25,0.75)}(t_1) = 2$. The line of a skyline tuple is called as the \textit{skyline line} (e.g., $l_1, l_2, l_3, l_4$ and $l_7$ in Figure \ref{fig-dual}). And $L(\mathrm{Sky}(D))$ denotes the list of skyline lines. Note that $l\in L$ is a skyline line, if there is no line in $L$ above $l$ for all $x\in [0,1]$. 

\begin{definition}[Convex Chain]
	\label{def-convex_chain}
	A size $r$ convex chain is a sequence of line segments, $\{s_1, \dots, s_r\}$, where $\forall i\in[r-1]$, $s_i$ and $s_{i+1}$ have a common endpoint and the slope of $s_i$ is less than that of $s_{i+1}$.
\end{definition} 

Based on Theorem \ref{t_skyline}, our goal is to find a size $r$ subset of $\mathrm{Sky}(D)$ that minimizes the rank-regret for $\mathbb{L}$. In dual space, the analog of such subset is a convex chain that begins on $y$-axis and ends on line $x=1$. Given a set $S\subseteq\mathrm{Sky}(D)$, the corresponding convex chain of $S$, denoted by $\mathrm{C}(S)$, is the upper envelop of skyline lines of $S$. 
Note that $S$ and $\mathrm{C}(S)$ may have different sizes. In Figure \ref{fig-dual}, $\mathrm{C}(\{t_2, t_3, t_4\})$ is the chain $\{(b, d), (d, g)\}$ (red curve). In the following, for notational convenience, $\mathrm{C}(S)$ is equivalently represented by a sequence of skyline lines, each of which contains a line segment in Definition \ref{def-convex_chain}. For example, $\mathrm{C}(\{t_1,t_3, t_7\})$ (green curve) is represented by $\{l_1, l_3, l_7\}$. The rank-regret of $S$ for $u=(c,1-c)$ is equal to one plus the number of lines in $L$ above $\mathrm{C}(S)$ for $x=c$, which is defined as the \textit{rank} of $\mathrm{C}(S)$ for $x=c$, denoted by $\nabla_{c}(\mathrm{C}(S))$. Obviously, $\nabla_{c}(\mathrm{C}(S))$ is the minimum rank of lines of $S$ for $x=c$, i.e., 
\vspace{-1ex}$$\nabla_{c}(\mathrm{C}(S)) = \min_{l\in L(S)}\nabla_{c}(l).\vspace{-1ex}$$
Note that the rank-regret of $S$ for $\mathbb{L}$ is equal to the maximum rank of $\mathrm{C}(S)$ for $x\in [0,1]$, abbreviated as 
\vspace{-1ex}$$\nabla_{[0,1]}(\mathrm{C}(S)) = \max_{c\in [0,1]} \nabla_{c}(\mathrm{C}(S)).\vspace{-1ex}$$ 
In Figure \ref{fig-dual}, the rank of chain $\{l_1, l_3, l_7\}$ for $x=0.25$ is 2, and its maximum rank for $x\in [0,1]$ is 3. 

\subsection{Algorithm Description}
\label{s-2d-ad}
We reformulate the RRM problem in 2D space and present the algorithm 2DRRM to find a size $r$ convex chain within $L(\mathrm{Sky}(D))$ to minimize the maximum rank, which gives an optimal solution for RRM in primal space. At a high-level, 2DRRM consists of two phases. 

In the first phase (line 1-2), 2DRRM sorts tuples of $D$ in descending order by the second attribute $A_2$. Then the corresponding lines, $l_1, l_2, \cdots , l_n$, is sorted by their intersections with line $x = 0$ from top to bottom. Meantime, it calculates $\mathrm{Sky}(D)$ and marks all skyline lines, $l_{g(1)}, l_{g(2)}, \cdots, l_{g(s)}$, where $g(i)$ is the initial position of $i$-th skyline line and $s$ is the number of skyline lines. Note that the skyline lines following index order are sorted by their slopes in strictly ascending order. In Figure \ref{fig-dual}, $l_7$ is the fifth skyline line and $g(5)=7$. Skyline lines $l_1, l_2, l_3, l_4, l_7$ are sorted by their slopes. The first phase takes $O(n\log n)$ time. 

In the second phase, 2DRRM makes a vertical line $\mathcal{L}$ move horizontally from $y$-axis to $x = 1$ (black dashed line in Figure \ref{fig-intersect}), stopping at the intersections of lines in $L$ (dots in Figure \ref{fig-intersect}). During the movement, 2DRRM traces out and evaluates the best convex chains encountered up to date, stored in a matrix $M$. When $\mathcal{L}$ reaches $x=1$, the best chain with a size bound $r$ is the optimal solution. Horizontal movement of $\mathcal{L}$ is achieved through two data structures: a sorted list of $L$ and a min-heap $H$, which maintains the unprocessed intersections. 

\textbf{Sorted List $\textbf{\textit{L}}$.} The lines in $L$ are always sorted by their intersections with $\mathcal{L}$ from top to bottom. In Figure \ref{fig-intersect}, when $\mathcal{L}$ reaches $x=0.25$ (blue dashed line), $L$ is the list $\{l_2, l_1, l_3, l_4, l_5, l_7, l_6\}$. Only after $\mathcal{L}$ passes through one of the intersections between lines in $L$, the sort order changes and $L$ is updated. And the order change is always between two adjacent lines in $L$. 
In Figure \ref{fig-intersect}, the intersection of line $l_1$ and $l_2$ is on the line $x=\frac{1}{9}$ (green dashed line). When $\mathcal{L}$ is prior to $x=\frac{1}{9}$, lines in $L$ maintain the original order, but just afterwards, the order of $l_1$ and $l_2$ flips.

\textbf{Min-Heap $\textbf{\textit{H}}$.} $H$ stores discovered but unprocessed intersections, sorted by x-coordinate in ascending order. The intersection of $l_i$ and $l_j$, denoted by $(l_i,l_j)$, is discovered exactly when $l_i$ and $l_j$ are immediate neighbors in $L$. In Figure \ref{fig-intersect}, when $\mathcal{L}$ passes through $x=\frac{1}{9}$, $l_1$ and $l_3$ become immediate neighbors and the intersection $(l_1,l_3)$ is inserted into $H$. To prevent repeated insertion, $H$ is implemented by a binary search tree. Before inserting an intersection, the algorithm first judges whether it is already in $H$.

\textbf{Matrix $\textbf{\textit{M}}$.} $M$ is a $s\times r$ matrix and maintains the best seen solutions. When $\mathcal{L}$ reaches $x=c$, the cell $M[i, j]$ stores the convex chain within $L(\mathrm{Sky}(D))$ that (i) ends in skyline line $l_{g(i)}$, (ii) contains at most $j$ segments, and (iii) minimizes the maximum rank for $x\in[0,c]$, stored in $M[i, j].rank$. In Figure \ref{fig-intersect}, when $\mathcal{L}$ reaches $x=0.25$, $M[2,2]$ contains the chain $\{l_1, l_2\}$ (blue curve) with $M[2,2].rank = \nabla_{[0,0.25]}(\{l_1, l_2\}) = 1$.

Afterwards, show the specific execution process. 2DRRM first initializes $H$ (line 4-6) and $M$ (line 7-8). For each pair of neighbors in $L$, if the intersection lies between $y$-axis and line $x=1$, push it into $H$. For each $i\in[s]$ and $j\in[r]$, $M[i, j]$ and $M[i, j].rank$ are initialized to the chain $\{l_{g(i)}\}$ and the rank of $l_{g(i)}$ in $L$. Next, 2DRRM processes the intersections in $H$. When $H$ is not empty, it pops the top intersection $(l_i,l_j)$ off $H$ (line 9-10), indicating $\mathcal{L}$ stops at $(l_i,l_j)$. Let $c$ be the $x$-coordinate of $(l_i,l_j)$. $l_i$ and $l_j$ are immediately adjacent in $L$. And $l_i$ is above $l_j$ for $x\in[0,c)$ and below $l_j$ for $x\in(c,1]$. Therefore, directly swap $l_i$ and $l_j$ in $L$ (line 11).  Then $l_i$ and $l_j$ have new neighbors $l_{p}$ and $l_{q}$ in $L$ respectively. Potentially, two new intersections are pushed into $H$ provided that they are between $\mathcal{L}$ and line $x=1$ (line 12-13). Updating $M$ is more complicated. When $\mathcal{L}$ passes through $(l_i,l_j)$, only the ranks of $l_i$ and $l_j$ change. Moreover, only the convex chains ends in line $l_i$ or $l_j$ may be updated. According to whether $l_i$ and $l_j$ are skyline lines, there are three cases: 
\begin{enumerate}[(1)]
	\item (Red and green dots in Figure \ref{fig-intersect}) $l_i$ is the $i'$-th skyline line, i.e., $g(i') = i$. For each $h\in [r]$, the rank of chain $M[i',h]$ is increased by one, and the maximum rank, i.e., $M[i',h].rank$, may also increase (line 14-16). (Red dots) Further, if $l_j$ is also a skyline line and $g(j') = j$, the rank of $M[j',h]$ is reduced by one, but its maximum rank remains unchanged. In addition, the old chain $M[j',h]$ may no longer be the best convex chain that meets the requirements. $M[i',h-1]$ and $l_j$ may form a better chain (line 17-19).
	\item (White dots) $l_j$ is a skyline line ($g(j') = j$), but $l_i$ is not. For all skyline lines, only the rank of $l_j$ changes. For each $h\in [r]$, the rank of $M[j',h]$ is reduced by one, but the maximum rank remains unchanged. No update.
	\item (Black dot) Both are not skyline lines. The ranks of skyline lines don't change. No update. 
\end{enumerate}
Finally, for all $i\in[s]$, 2DRRM finds the cell $M[i,r]$ with the lowest $M[i,r].rank$ (line 20). For each skyline line in $M[i,r]$, it adds corresponding tuple to the result set $S$ (line 21).

Briefly illustrate the process of handling intersections. Assume $r = 2$ and $D$ contains only $t_1$, $t_2$ and $t_3$ in Table \ref{tab_example}. $L$ and $H$ are first set to $\{l_1, l_2, l_3\}$ and $\{(l_1, l_2), (l_2, l_3)\}$ respectively. And $M$ is initialized as shown in the second column of Table \ref{tab_m} where "-" means no update. Then 2DRRM takes the first intersection $(l_1, l_2)$ from $H$, and $L$ is changed into $\{l_2,l_1,l_3\}$. The new intersection $(l_1,l_3)$ is inserted into $H$. Since $l_1$ is the $1$-st skyline line (i.e., $g(1)=1$), $M[1,2].rank$ is updated as $2$. Further, $l_2$ is the $2$-nd skyline line, and $M[1,1]$ with $l_2$ forms a better convex chain, i.e., $M[2,2].rank>M[1,1].rank$. Then $M[2,2]$ and $M[2,2].rank$ are modified to $\{l_1, l_2\}$ and $1$. Subsequently, $M[1,1].rank$ is changed into $2$. After processing $(l_1,l_3)$ and $(l_2,l_3)$ in turn, 2DRRM returns $\{t_1, t_2\}$ or $\{t_1, t_3\}$.

\begin{figure}[!t]
	\removelatexerror
	\begin{algorithm}[H]
		\SetKwBlock{DoWhile}{Do}{end}
		\SetKwProg{Def}{def}{}{end}
		\caption{2DRRM}
		\KwIn{$D$, $r$;} 
		\KwOut{a size $r$ set $S$ that minimizes the rank-regret;}
		Sort tuples in $D$ and calculate $\mathrm{Sky}(D)$\;
		Transform tuples in $D$ to lines in $L$\;
		$S$ is an empty set, $H$ is a min-heap and $M$ is a matrix\;
		\ForEach{$i\in [n-1]$}{
			\If{$(l_i, l_{i+1})$ lies between $x=0$ and $x=1$}{
				push $(l_i, l_{i+1})$ into $H$\;
			}
		}
		\ForEach{$i\in[s]$ and $j\in[r]$}{
			$M[i,j]$ = $\{l_{g(i)}\}$, $M[i,j].rank$ = $Rank(l_{g(i)})$\;
		}
		\While{$H$ is not empty}{
			Pop top intersection $(l_i, l_j)$ off $H$\;
			Swap $l_i$ and $l_j$ in $L$; \tcp{$l_{p}$ and $l_{q}$ are new neighbors of $l_i$ and $l_j$ in $L$}
			\textbf{if} $(l_q, l_j)$ lies between $\mathcal{L}$ and $x=1$, push it into $H$\;
			\textbf{if} $(l_i, l_p)$ lies between $\mathcal{L}$ and $x=1$, push it into $H$\;
			\If{$l_i$ is a skyline line (and $i=g(i')$)}{
				\For{$h = r, r-1, \cdots, 2, 1$}{
					$M[i',h].rank$ = $\max$($M[i',h].rank$, $Rank(l_i)$)\;
					\If{$l_j$ is a skyline line (and $j=g(j')$) and $M[j',h].rank > M[i',h-1].rank$}{
						$M[j',h].rank = M[i',h-1].rank$\;
						$M[j',h] = M[i',h-1]$ suffixed with $l_j$\;
					}
				}
			}
		}
		Find $M[i,r]$ for $i\in[s]$ with lowest $M[i,r].rank$\;
		Add corresponding tuples of lines in $M[i,r]$ to $S$\;
		\Return $S$\;
	\end{algorithm}
	\vspace{-3ex} 
\end{figure}

\begin{theorem}
	In 2D space, 2DRRM returns an optimal solution for the RRM problem. 
\end{theorem}
\begin{IEEEproof}
	Based on Theorem \ref{t_skyline}, there must be a size $r$ subset of $\mathrm{Sky}(D)$ that is the optimal solution for the RRM problem. Let $k^*$ be its rank-regret for $\mathbb{L}$. Let $\mathrm{C}^*$ be its corresponding convex chain, denoted by a sequence of skyline lines $\{l_{g(i_1)}, l_{g(i_2)}, \cdots, l_{g(i_r)}\}$, where $1\le i_1<i_2<\cdots<i_r\le s$ and $l_{g(i)}$ is the $i$-th skyline line. We prove that for any integer $j\in [r-1]$, when $\mathcal{L}$ passes intersection $(l_{g(i_j)}, l_{g(i_{j+1})})$ and $M$ has been updated, $M[i_{j+1},j+1].rank\le k^*$. 
	
	We first prove that it holds for $j=1$. Let $c_h$ be the x-coordinate of intersection $(l_{g(i_h)},l_{g(i_{h+1})})$. Since the maximum rank of convex chain $\mathrm{C}^*$ is $k^*$, the rank of $l_{g(i_1)}$ is at most $k^*$ for $x\in[0,c_1]$. Further, $M[i_1,1]$ is $\{l_{g(i_1)}\}$. Thence, before $\mathcal{L}$ reaches $(l_{g(i_1)}, l_{g(i_2)})$, $M[i_1,1].rank$ is at most $k^*$. When $\mathcal{L}$ passes $(l_{g(i_1)}, l_{g(i_2)})$, $M[i_2,2].rank$ is updated as the smaller of its old value and unupdated $M[i_1,1].rank$. Thus $M[i_2,2].rank\le k^*$.
	
	We prove if it holds for $j=h$ $(h\in [r-2])$, then it also holds for $j=h+1$. Assume the maximum rank of $M[i_{h+1},h+1]$ for $x\in[0, c_h]$ is at most $k^*$. When $\mathcal{L}$ moves from line $x=c_h$ to line $x=c_{h+1}$, the cell $M[i_{h+1},h+1]$ may be updated. However, the rank of $l_{g(i_{h+1})}$ is at most $k^*$ for $x\in[c_h, c_{h+1}]$. Therefore, the maximum rank of $M[i_{h+1},h+1]$ for $x\in[0, c_{h+1}]$, i.e., the value of $M[i_{h+1},h+1].rank$ before $\mathcal{L}$ passes $(l_{g(i_{h+1})}, l_{g(i_{h+2})})$, still does not exceed $k^*$. Same as before, $M[i_{h+2},h+2].rank\le k^*$.
	
	Thence, when $\mathcal{L}$ passes ($l_{g(i_{r-1})}$, $l_{g(i_r)}$) and $M$ has been updated, the maximum rank of $M[i_r,r]$ for $x\in[0,c_{r-1}]$ is at most $k^*$. Further, the rank of $l_{g(i_r)}$ is at most $k^*$ for $x\in[c_{r-1},1]$. Therefore, when the algorithm terminates, the maximum rank of $M[i_r,r]$ for $x\in[0,1]$ is at most $k^*$. And the corresponding set of skyline tuples has a rank-regret of $k^*$ for $\mathbb{L}$. 
\end{IEEEproof}

\begin{theorem}
	2DRRM takes $O(n^2\log n)$ time and uses $O(n^2)$ space.
\end{theorem}
\begin{IEEEproof}
	The first phase (line 1-2) takes $O(n\log n)$ time. It takes $O(n\log n+sr)$ time to initialize $H$ and $M$ (line 3-8). There are at most $n(n-1)/2$ intersections. It takes $O(\log n)$ time to pop or push an intersection to $H$. It takes $O(1)$ time to swap two adjacent lines in $L$. If the intersection corresponds to case (2) or case (3), the algorithm dose not perform any operation in line 14-19. Otherwise, it has to update $O(r)$ cells in $M$. There are at most $sn$ intersections that fit case (1). Therefore, updates to $L$, $H$ and $M$ for all intersections take $O(n^2\log n+snr)$ time, which dominates the running time of the algorithm. 
	
	The space comes from the min-heap $H$ and the matrix $M$. There are at most $O(n^2)$ intersections in $H$ and $O(sr)$ cells in $M$.
\end{IEEEproof}

2DRRM can be easily adopted for RRR\cite{asudeh2019rrr} by a binary search with an additional $\log n$ factor in the running time, and returns the optimal solution. 2DRRM is inspired by the 2D algorithm proposed by Chester et al. \cite{chester2014computing} for the kRMS problem, but there are many differences between them. First, the dual transformation used in 2DRRM is more practical and understandable. Secondly, based on Theorem \ref{t_skyline}, 2DRRM uses only the skyline tuples as candidates, not all tuples in $D$. Thirdly, 2DRRM can be generalized to the restricted function space, shown in the following subsection. Last but not least, the goal of 2DRRM is to minimize rank-regret, rather than $k$-regret-ratio \cite{chester2014computing}. And there is no need to pre-calculate the top-$k$ rank contour \cite{chester2014computing}.

\subsection{Modifications for RRRM}
Through some modifications, 2DRRM can return the optimal solution for the RRRM problem in 2D space. For RRRM, utility vectors are bounded in a specific convex region $\mathbb{U}\subseteq\mathbb{L}$. Earlier, it is assumed that each $u\in\mathbb{L}$ is normalized such that $u[1] + u[2] = 1$. Similarly, normalize the vectors in $\mathbb{U}$. Then they lie on a line segment 
\vspace{-1ex} $$\mathbb{P}=\{(u[1],1-u[1])\ |\ \exists u'\in\mathbb{U}, u[1]=\frac{u'[1]}{u'[1]+u'[2]}\}. \vspace{-1ex} $$ 
For example, in Figure \ref{fig-render}, $\mathbb{U}$ is the shaded region, and then $\mathbb{P}$ is the red line segment. Let $c_0$ and $c_1$ be the $x$-coordinates of two endpoints. The above conversion process is called \textit{rendering the scene} \cite{de1997computational} in computational geometry and takes $O(1)$ time. 

In addition to the conversion process, 2DRRM requires three additional modifications. Firstly, it focuses on the subsets of $\mathrm{Sky}_\mathbb{U}(D)$ to find the solution of RRRM, rather than $\mathrm{Sky}(D)$. In 2D space, $\mathrm{Sky}_\mathbb{U}(D)$ can be computed in $O(n\log n)$ time \cite{liu2021eclipse}. Secondly, in the first phase, tuples are sorted based on utilities w.r.t. $u=(c_0, 1-c_0)$, rather than the values on $A_2$. Thirdly, $\mathcal{L}$ moves horizontally from line $x=c_0$ to line $x = c_1$, rather than from $y$-axis to $x=1$. Therefore, $H$ only stores the intersections that lie between $\mathcal{L}$ and $x = c_1$, and $M$ maintains the convex chains that minimize the maximum rank for $x\in[c_0,c]$ ($\mathcal{L}$ is line $x=c$ and $c\in[c_0,c_1]$).

%% file: file/HD.tex
\section{Algorithm in HD}
\label{s_hd}
In this section, discuss the problems in HD space. Both RRM and RRRM are NP-hard for $d\ge 3$. Consider designing an approximation algorithm. 

\begin{figure}[tb]
	\centering
	\begin{minipage}[t]{0.43\linewidth}
		\centering
		\includegraphics[width=1\linewidth]{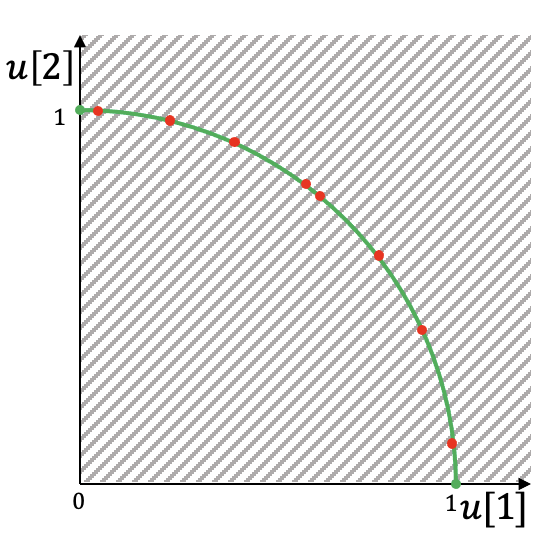} 
		\caption{$\mathbb{L}$ (Shaded), $\mathbb{S}$ (Green), and $\mathbb{D}$ (Red) in 2D space.}
		\label{fig-sphere}
	\end{minipage}
	\hspace{0.03\linewidth}
	\begin{minipage}[t]{0.43\linewidth}
		\centering
		\includegraphics[width=1\linewidth]{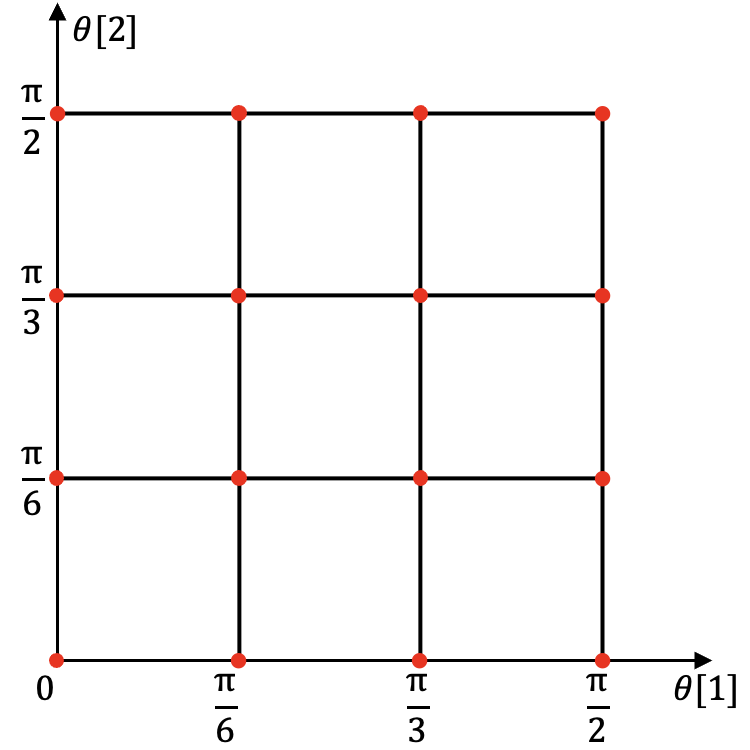} 
		\caption{Illustration of space partition in polar when $d = 3$ and $\gamma = 3$}
		\label{fig-polar}
	\end{minipage}
	\vspace{-3ex} 
\end{figure}

In HD space, the geometric shapes become complex and the problem conversion in 2DRRM is inefficient. With the help of sampling and the polar coordinate system, we take an alternative route for solving RRM by transforming the full space $\mathbb{L}$ into a discrete set $\mathbb{D}$. From two different perspectives, it is proved that if a representative set has a small rank-regret for $\mathbb{D}$, then it is almost the same for $\mathbb{L}$. Based on the finite size of $\mathbb{D}$, consider finding a given size set to minimize the rank-regret w.r.t. $\mathbb{D}$, which is an approximate solution for RRM. The problem is converted into linear number of set-cover\cite{bronnimann1995almost} instances, and then the algorithm HDRRM is proposed, which introduces a double approximation guarantee for rank-regret. Finally, it is shown that HDRRM is suitable for RRRM. 

\subsection{Algorithm Preparation}
\label{subs_guarrantee}
Without loss of generality, assume that each $u$ in $\mathbb{L}$ is normalized such that $\Vert u\Vert = 1$ in this section. Then the vectors lie on the surface $\mathbb{S}$ of a sphere with radius $1$ centered at the origin in the $d$-dimensional non-negative orthant, i.e., 
\vspace{-1ex} $$\mathbb{S}=\{u\ |\ u\in\mathbb{R}^d_+ \land \Vert u\Vert = 1 \}. \vspace{-1ex} $$ 
Note that for any $S\subseteq D$, $\nabla_{\mathbb{L}}(S) = \nabla_{\mathbb{S}}(S)$. There is an infinite number of vectors in $\mathbb{S}$. To make "infinite" become "finite", construct a discrete set $\mathbb{D}\subseteq \mathbb{S}$, e.g., Figure \ref{fig-sphere} shows $\mathbb{L}$ (shaded region), $\mathbb{S}$ (green quarter arc) and $\mathbb{D}$ (set of red dots) in 2D space. $\mathbb{D}$ contains two parts: $\mathbb{D}_a$, a size $m$ set of samples generated from $\mathbb{S}$ by a $d$-dimensional uniform distribution, to estimate the area of some subregions of $\mathbb{S}$; $\mathbb{D}_b$, a size $(\gamma+1)^{d-1}$ set obtained by discretization in polar coordinate space with a parameter $\gamma$, to guarantee that for any $u\in\mathbb{S}$, there is a $v\in\mathbb{D}_b$ close to $u$. Note that $\mathbb{D} = \mathbb{D}_a \cup \mathbb{D}_b$. From two different perspectives, it is proved that if a set has a small rank-regret for $\mathbb{D}$, then it is approximately the same for $\mathbb{S}$. 

Inspired by Asudeh et al. \cite{asudeh2017efficient}, $\mathbb{D}_b$ is obtained with the help of polar coordinate system. Each vector $(u[1],\cdots,u[d])$ in $\mathbb{S}$ corresponds to a $(d-1)$-dimensional angle vector $(\theta[1], \dots, \theta[d-1])$ in polar (since $\Vert u\Vert=1$, by default, the magnitude is 1), where $\theta[0]$ is set to $0$ and $\forall i\in[d]$,  
\vspace{-1ex} $$u[i]=\sin(\theta[d-1])\sin(\theta[d-2])\times\cdots\times\sin(\theta[i])\cos(\theta[i-1]). \vspace{-1ex} $$ 
In polar, divide the range of each angle dimension, i.e., $[0,\frac{\pi}{2}]$, into $\gamma$ equal-width segments. The width of each segment is $\frac{\pi}{2\gamma}$. Keep the $\gamma+1$ boundaries of segments on each dimension, and then get a size $(\gamma+1)^{d-1}$ set of angle vectors, i.e., 
\vspace{-1ex} $$\{(\frac{z_1\pi}{2\gamma},\cdots,\frac{z_{d-1}\pi}{2\gamma})\ |\ \forall i\in[d-1], z_i\in\{0,1,\dots,\gamma\}\}. \vspace{-1ex} $$ 
As a example, Figure \ref{fig-polar} illustrates the above set of angle vectors (red dots) in polar when $d = 3$ and $\gamma = 3$.
Based on polar coordinate conversion, transform the set to a subset $\mathbb{D}_b$ of $\mathbb{S}$. 

Next, prove that the discrete set $\mathbb{D}$ can approximate $\mathbb{S}$. Given a tuple $t\in D$ and an integer $k\in[n]$, let $V_{\mathbb{S}, k}(t)$ denote the set of vectors in $\mathbb{S}$ for which $t$ is ranked in top-$k$ of $D$, i.e., 
\vspace{-1ex} $$V_{\mathbb{S}, k}(t) = \{u\in\mathbb{S}\ |\ w(u,t)\ge w_k(u,D)\}. \vspace{-1ex} $$ 
If $k$ is obvious from the context, use $V_\mathbb{S}(t)$ to denote $V_{\mathbb{S}, k}(t)$. Given a set $S\subseteq D$, $\cup_{t\in S} V_\mathbb{S}(t)$ is a partial spherical surface of $\mathbb{S}$. Intuitively, if there are more vectors in $\cup_{t\in S} V_\mathbb{S}(t)$, $S$ is more representative. But the number of vectors in $\cup_{t\in S} V_\mathbb{S}(t)$ is uncountable. Define $k$-$\mathit{ratio}$ of $S$, denoted by $\mathrm{Rat}_k(S)$, as the ratio of surface area of $\cup_{t\in S} V_\mathbb{S}(t)$ to that of $\mathbb{S}$, i.e., 
\vspace{-1ex} $$\mathrm{Rat}_k(S) = \frac{\mathrm{Area}(\cup_{t\in S} V_\mathbb{S}(t))}{\mathrm{Area}(\mathbb{S})} \vspace{-1ex} $$ 
where $\mathrm{Area}(\cdot)$ indicates the surface area. $\mathrm{Rat}_k(S)$ is the proportion of vectors in $\mathbb{S}$, for which $S$ has a rank-regret at most $k$. Further, we have the following lemma. 
\begin{lemma}
	\label{l_equivalent}
	For any $S\subseteq D$, $\nabla_\mathbb{S}(S)\le k$, iff $\mathrm{Rat}_k(S) = 1$. 
\end{lemma}
\begin{IEEEproof} 
	Assume $S$ has a rank-regret no more than $k$ for $\mathbb{S}$. Then $\forall u\in\mathbb{S}$, there is a tuple $t\in S$ such that $w(u,t)\ge w_k(u,D)$, i.e., $u\in V_\mathbb{S}(t)$. Thus we have $\cup_{t\in S} V_\mathbb{S}(t)=\mathbb{S}$ and $\mathrm{Rat}_k(S) = 1$. 
	
	Assume $\mathrm{Rat}_k(S) = 1$. Then $\cup_{t\in S} V_\mathbb{S}(t) = \mathbb{S}$ and $\forall u\in\mathbb{S}$, there is a tuple $t\in S$ such that $u\in V_\mathbb{S}(t)$, which means that $\nabla_{u}(S)\le k$. Thus we have $\nabla_\mathbb{S}(S)\le k$.
\end{IEEEproof}

Following the setting in \cite{xie2020being}, assume each vector in $\mathbb{S}$ has the same probability of being used by a user, i.e., for any user, his/her utility vector is a random variable and obeys a uniform distribution on $\mathbb{S}$. We relax the assumption later and show that the analysis can be applied for any other distribution. Under the assumption, for any user, $\mathrm{Rat}_k(S)$ is the probability that $S$ contains a top-$k$ tuple w.r.t. his/her utility function. 

Further, $V_{\mathbb{D}_a,k}(t)$, abbreviated as $V_{\mathbb{D}_a}(t)$, is the set of vectors in $\mathbb{D}_a$ based on which $t$ is ranked in the top-$k$ of $D$, i.e., 
\vspace{-1ex} $$V_{\mathbb{D}_a,k}(t)=\{u\in\mathbb{D}_a | w(u,t)\ge w_k(u,D)\}. \vspace{-1ex}$$ 
Since $\mathbb{D}_a$ is a size $m$ unbiased sample of $\mathbb{S}$, $\frac{|\cup_{t\in S} V_{\mathbb{D}_a}(t)|}{m}$ is an unbiased estimate of $\mathrm{Rat}_k(S)$. Due to $\mathbb{D}_a\subseteq\mathbb{D}$, if $\nabla_{\mathbb{D}}(S)\le k$, then $\nabla_{\mathbb{D}_a}(S)\le k$ and $\frac{|\cup_{t\in S} V_{\mathbb{D}_a}(t)|}{m}=1$. And we have the following theorem. 
\begin{theorem}
	\label{t_guarantee_2}
	Given a collection $\mathcal{S}\subseteq 2^D$, a set $\mathcal{K}\subseteq [n]$ and a sample size $m=O(\frac{\ln|\mathcal{S}|+\ln n}{\varepsilon^2})$, with probability at least $1-\frac{1}{n}$, $\forall S\in\mathcal{S}$ and $\forall k\in\mathcal{K}$, if $\nabla_{\mathbb{D}}(S)\le k$, then
	\vspace{-1ex} $$ \mathrm{Rat}_k(S)\ge1-\varepsilon. \vspace{-1ex}$$ 
\end{theorem}
\begin{IEEEproof}
	Let $S$ be a set in $\mathcal{S}$ and $k$ be an integer in $\mathcal{K}$. For each $i\in[m]$, let $u_i$ be the $i$-th vector in $\mathbb{D}_a$ and $X_i$ be a random variable where $X_i= 1$ if $u_i\in \cup_{t\in S} V_\mathbb{S}(t)$, otherwise $X_i= 0$. Then probability $\mathrm{Pr}(X_i= 1)$ is equal to $\mathrm{Rat}_k(S)$. And $|\cup_{t\in S} V_{\mathbb{D}_a}(t)| = \sum_{i=1}^{m}X_i$. According to the well-known Chernoff-Hoeffding Inequality \cite{phillips2012chernoff}, we know that with a probability at most $\frac{1}{|\mathcal{S}||\mathcal{K}|n}$, 
	\begin{equation} \frac{|\cup_{t\in S} V_{\mathbb{D}_a}(t)|}{m}-\mathrm{Rat}_k(S) >  \sqrt{\frac{\ln|\mathcal{S}|+\ln|\mathcal{K}|+\ln n}{2m}}. \label{equ_1} \end{equation}
	Therefore, with a probability at least $1-\frac{1}{n}$, $\forall S\in\mathcal{S}$ and $\forall k\in\mathcal{K}$, inequality (\ref{equ_1}) does not hold. If $\nabla_{\mathbb{D}}(S)\le k$, then $\frac{|\cup_{t\in S} V_{\mathbb{D}_a}(t)|}{m}=1$. Set $m$ to $\frac{\ln|\mathcal{S}|+\ln|\mathcal{K}|+\ln n}{2\varepsilon^2}$, and then the theorem is proved. 
\end{IEEEproof}

Theorem \ref{t_guarantee_2} shows that if a set $S$ satisfies $\nabla_{\mathbb{D}}\le k$, the value of $\mathrm{Rat}_k(S)$ is close to $1$, i.e., for any user, with high probability, $S$ has a rank-regret at most $k$ w.r.t. his/her utility function. In other words, if a set has a small rank-regret for $\mathbb{D}$, it is approximately the same for $\mathbb{S}$. Next, prove it from another perspective.

\begin{theorem}
	\label{t_guarantee_1}
	Given $k\in[n]$, a set $S\subseteq D$ and a discretization parameter $\gamma = O(\frac{1}{\epsilon})$, if $\nabla_\mathbb{D}(S)\le k$ and $B\subseteq S$, then $\forall u\in\mathbb{S}$, 
	\vspace{-1ex} $$w(u,S)\ge (1-\epsilon)w_k(u,D)\vspace{-1ex}, $$
	where 
	$B$ is the basis of $D$. 
\end{theorem}
\begin{IEEEproof}
	Given a utility vector $u\in\mathbb{S}$, we have to prove that there is a tuple $t\in S$ such that $w(u,t)\ge (1-\epsilon)w_k(u,D)$. Discuss two cases for $u$ separately. Firstly, consider the case that $w_k(u,D)\le \frac{1}{(1-\epsilon)\sqrt{d}}$. Given a utility vector $u\in\mathbb{S}$, since $\Vert u\Vert=1 $, there must be an attribute $A_i$ such that $u[i]\ge1/\sqrt{d}$. Based on the definition of basis, for $A_i$, there is also a tuple $t$ in $B$ such that $t[i] = 1$. And then $w(u, t)\ge 1/\sqrt{d}$. Due to $B\subseteq S$, the theorem holds at this time. 
	
	Secondly, consider the case that $w_k(u,D)> \frac{1}{(1-\epsilon)\sqrt{d}}$. As shown in \cite{asudeh2017efficient}, there is a utility vector $v\in\mathbb{D}_b$ such that
	\begin{align*} \Vert u-v\Vert &\le  \sqrt{\frac{1}{2}-\frac{1}{2}\cos^{d-1}(\frac{\pi}{2\gamma})}\\ &=  \sqrt{\frac{1}{2}-\frac{1}{2}(1-2\sin^2\frac{\pi}{4\gamma})^{d-1}}\\ &\le \sqrt{d-1}\sin\frac{\pi}{4\gamma} \le \sqrt{d-1}\frac{\pi}{4\gamma} = \sigma \end{align*}
	The range of tuples in $D$ on each attribute is normalized to $[0,1]$. Thus $\forall t\in D$, $\Vert t\Vert\le\sqrt{d}$ and $$ |w(u,t)-w(v,t)|=\langle u-v, t\rangle \le \Vert u-v\Vert\cdot\Vert t\Vert\le\sigma\sqrt{d}. $$ For each $t$ in the top-$k$ of $D$ w.r.t. $u$, we have $$ w(v,t)\ge w(u,t)-\sigma\sqrt{d}\ge w_k(u,D)-\sigma\sqrt{d}. $$ For utility vector $v$, there are at least $k$ tuples in $D$ whose utilities are no less than $w_k(u,D)-\sigma\sqrt{d}$. It means that $w_k(v,D)$, i.e., the $k$-th highest utility w.r.t. $v$, is at least $w_k(u,D) - \sigma\sqrt{d}$. Since $\mathbb{D}_b \subseteq \mathbb{D}$, $\nabla_{\mathbb{D}_b}(S)\le\nabla_\mathbb{D}(S)\le k$ and there is a tuple $t'$ in $S$ such that $w(v,t') \ge w_k(v,D)$. Then we have $$ w(u,t')\ge w(v,t')-\sigma\sqrt{d}\ge w_k(u,D)-2\sigma\sqrt{d}. $$ Due to $w_k(u,D)> \frac{1}{(1-\epsilon)\sqrt{d}}$, we have $$ w(u,t') > (1-2d\sigma(1-\epsilon))w_k(u,D) > (1-2d\sigma)w_k(u,D). $$ Let $\gamma$ be $\frac{d\sqrt{d-1}\pi}{2\epsilon}$. Then $w(u,t')> (1-\epsilon)w_k(u,D)$ and the theorem holds at this time. 
\end{IEEEproof}
Note that $S$ has a rank-regret at most $k$ for $\mathbb{S}$, if $\forall u\in\mathbb{S}$, $w(u,S)\ge w_k(u,D)$. Theorem \ref{t_guarantee_1} shows from another perspective that $\mathbb{D}$ can approximate $\mathbb{S}$.

\subsection{Algorithm for RRM}
Based on Theorem \ref{t_guarantee_2} and \ref{t_guarantee_1}, we propose an approximation algorithm HDRRM to find a size $r$ superset of basis $B$ with the minimum rank-regret for $\mathbb{D}$, which approaches the optimal solution of RRM. The rank-regret of a subset of $D$ is in set $[n]$. Consider each value in $[n]$ as a possible rank-regret threshold of the following problem.

\begin{definition}[MS Problem] 
	Given a dataset $D$, a threshold $k\in[n]$, the basis $B$ of $D$ and the utility vector set $\mathbb{D}$, find the minimum superset of $B$ with a rank-regret at most $k$ for $\mathbb{D}$, i.e., return a set 
	\vspace{-1ex} $$ Q^* = \mathop{\arg\min}\limits_{B\subseteq Q\subseteq D: \nabla_\mathbb{D}(Q)\le k} |Q|. $$ 
\end{definition}

Assume that there is a solver for MS. Given the size bound $r$, a naive algorithm for RRM applies binary search to explore the values of $k$ in $[n]$ and for each value, calls the solver. 

\subsubsection{Approximate Solver for MS}
\begin{figure}[!t]
	\removelatexerror
	\begin{algorithm}[H]
		\SetKwBlock{DoWhile}{Do}{end}
		\SetKwProg{Def}{def}{}{end}
		\caption{ASMS}
		\KwIn{$D$, $k$, $B$, $\mathbb{D}$;} 
		\KwOut{a superset $Q$ of $B$ with $\nabla_\mathbb{D}(Q)\le k$;}
		$Q=B$, $\mathcal{V}=\emptyset$, $\mathbb{D}_k=\emptyset$, and $\forall t\in D$, $V_{\mathbb{D}_k}(t)=\emptyset$\;
		\ForEach{$u\in\mathbb{D}$}{
			Compute $\Phi_k(u,D)$\;
			\If{$\Phi_k(u,D)$ contains no boundary tuple}{
				Add $u$ to $\mathbb{D}_k$\;
				$\forall t\in\Phi_k(u,D)$, add $u$ to $V_{\mathbb{D}_k}(t)$\;
			}
		}
		$\forall t\in D$, add non-empty $V_{\mathbb{D}_k}(t)$ to $\mathcal{V}$\;
		$\mathcal{C}$ = Set-Cover($\mathbb{D}_k$, $\mathcal{V}$)\;
		Add corresponding tuples of $\mathcal{C}$ to $Q$\;
		\Return $Q$\;
	\end{algorithm}
	\vspace{-3ex} 
\end{figure}

First, convert the MS problem into a set-cover instance \cite{bronnimann1995almost}. Note that given a set $Q\subseteq D$, $\nabla_\mathbb{D}(Q)\le k$, if $\cup_{t\in Q} V_{\mathbb{D},k}(t) = \mathbb{D}$, where $V_{\mathbb{D},k}(t)$ is the set of vectors in $\mathbb{D}$ for which $t$ is ranked in top-$k$ of $D$. $V_{\mathbb{D},k}(t)$ can be viewed as the set of vectors "covered" by $t$, and MS aims to find a small tuple set that "covers" all vectors in $\mathbb{D}$. Since the target set must contain all boundary tuples, there is no need to take vectors "covered" by them into consideration. Let $\mathbb{D}_k$ be the set of vectors in $\mathbb{D}$ "uncovered" by boundary tuples, i.e., 
\vspace{-1ex} $$\mathbb{D}_k = \mathbb{D}\backslash(\cup_{t\in B} V_{\mathbb{D},k}(t)). \vspace{-1ex} $$ 
Similarly, $V_{\mathbb{D}_k}(t)$ denotes the set of vectors in $\mathbb{D}_k$ "covered" by tuple $t$, i.e., 
\vspace{-1ex} $$V_{\mathbb{D}_k}(t) = \{u\in\mathbb{D}_k\ |\ w(u,t)\ge w_k(u,D)\}. \vspace{-1ex} $$ 
\begin{lemma}
	\label{l_Qk}
	For any $Q\subseteq D$ with $B\subseteq Q$, $\nabla_\mathbb{D}(Q)\le k$, iff $\cup_{t\in Q} V_{\mathbb{D}_k}(t)=\mathbb{D}_k$. 
\end{lemma}
\begin{IEEEproof} 
	Assume $Q$ has a rank-regret no more than $k$ for $\mathbb{D}$. Then for each vector $u$ in $\mathbb{D}_k\subseteq\mathbb{D}$, there is a tuple $t\in Q$ such that $w(u,t)\ge w_k(u,D)$, i.e., $u\in V_{\mathbb{D}_k}(t)$. Thus we have $\cup_{t\in Q} V_{\mathbb{D}_k}(t)=\mathbb{D}_k$. 
	
	Assume $\cup_{t\in Q} V_{\mathbb{D}_k}(t) = \mathbb{D}_k$. Since $V_{\mathbb{D}_k}(t) \subseteq V_\mathbb{D}(t)$, we have $\mathbb{D}_k \subseteq \cup_{t\in Q} V_\mathbb{D}(t)$. Due to $B\subseteq Q$, $\cup_{t\in B}V_\mathbb{D}(t) \subseteq \cup_{t\in Q} V_\mathbb{D}(t)$. Further, we know that $\mathbb{D}_k\cup(\cup_{t\in B} V_\mathbb{D}(t)) = \mathbb{D}$. Thence $\mathbb{D} \subseteq \cup_{t\in Q}V_\mathbb{D}(t) \subseteq \mathbb{D}$, and $\nabla_\mathbb{D}(Q)\le k$. 
\end{IEEEproof}

Lemma \ref{l_Qk} indicates that the goal set just needs to "cover" the vectors in $\mathbb{D}_k$. Then MS is formulated as a set-cover\cite{bronnimann1995almost} instance, and its solution is obtained by running a set-cover algorithm on the set system $\Sigma= (\mathbb{D}_k, \mathcal{V})$, where 
\vspace{-1ex} $$\mathcal{V} = \{V_{\mathbb{D}_k}(t)\ |\ t\in D\land V_{\mathbb{D}_k}(t)\ne\emptyset\}. \vspace{-1ex} $$ 
It aims to find the minimum set-cover of $\Sigma$, where a collection $\mathcal{C}\subseteq\mathcal{V}$ is a set-cover if $\cup_{V\in\mathcal{C}}V = \mathbb{D}_k$. A set-cover $\mathcal{C}$ of $\Sigma$ corresponds to a set $S\subseteq D$ such that $\cup_{t\in S} V_{\mathbb{D}_k}(t)=\mathbb{D}_k$. And $Q=S\cup B$ has a rank-regret at most $k$ for $\mathbb{D}$. 

Based on the above analysis, we propose an approximate solver for MS, called ASMS, whose pseudo-code is shown in Algorithm 2. In ASMS, Set-Cover($\cdot, \cdot$) (line 8) is a naive greedy algorithm proposed by Chvatal \cite{chvatal1979greedy} for the set-cover problem. Starting from $\mathcal{C} = \emptyset$, it always adds the set in $\mathcal{V}$ that contains the largest number of uncovered vectors, to $\mathcal{C}$ at each iteration until $\cup_{V\in \mathcal{C}}V = \mathbb{D}_k$. Theoretically, Set-Cover($\cdot, \cdot$) has a time complexity of $O(|\mathbb{D}_k|\cdot|\mathcal{V}|)$, and its result achieves an approximation ratio of $1+\ln|\mathbb{D}_k|$ on size. 

\begin{theorem} 
	\label{t_RRR_time} 
	ASMS is an $O(|\mathbb{D}|(nd+k\log k))$ time solver. 
\end{theorem}
\begin{IEEEproof} 
	Since there are $n$ tuples in $D$, line 1 takes $O(n)$ time. For each $u\in\mathbb{D}$, $\Phi_k(u,D)$ is computed in $O(nd+k\log k)$ time. Thus line 2-7 takes $O(|\mathbb{D}|(nd+k\log k))$ time totally. There are at most $|\mathbb{D}|$ vectors in $\mathbb{D}_k$ and $n$ sets in $\mathcal{V}$, and then the time complexity of the invoked set-cover algorithm is $O(|\mathbb{D}|n)$. Therefore, ASMS takes $O(|\mathbb{D}|(nd+k\log k))$ time. 
\end{IEEEproof}

\subsubsection{Improved Binary Search}
\begin{figure}[!t]
\removelatexerror
\begin{algorithm}[H]
	\SetKwBlock{DoWhile}{Do}{end}
	\SetKwProg{Def}{def}{}{end}
	\caption{HDRRM}
	\KwIn{$D$, $r$, $\gamma$, $m$;} 
	\KwOut{a representative set $R\subseteq D$ no larger than $r$;}
	Construct the vector set $\mathbb{D}$ and compute the basis $B$\;
	$k = 1$\;
	\While{$k\le n$}{
		$Q$ = ASMS($D$, $k$, $B$, $\mathbb{D}$)\;
		\textbf{if} $|Q|\le r$, \textbf{then} $R = Q$ and \textbf{break}\;
		$k=2k$\;
	}
	$low = k/2+1$, $high = k$\;
	\While{$low<high$}{
		$k=\lfloor(low+high)/2\rfloor$\;
		$Q$ = ASMS($D$, $k$, $B$, $\mathbb{D}$)\;
		\textbf{if} $|Q|\le r$, \textbf{then} $high=k$ and $R = Q$\;
		\textbf{else} $low = k+1$\;
	}
	\Return $R$\;
\end{algorithm}
\vspace{-3ex} 
\end{figure}

Theorem \ref{t_RRR_time} shows that ASMS has a time complexity related to the threshold $k$. To make the value of $k$ as small as possible in ASMS, HDRRM turns the original binary search into two stages. In the first stage, it repeatedly doubles the value of $k$ and calls ASMS, until a set no larger than $r$ is obtained (line 3-6). Then the search space is $\{k/2+1, k/2+2, \dots, k\}$ (line 7), on which HDRRM performs a binary search in the second stage (line 8-12). 

\subsubsection{Theoretical Results}
\begin{theorem}
	\label{t_RRR_quality}
	Given a integer $k\ge1$, ASMS returns a set $Q$ such that $\nabla_\mathbb{D}(Q)\le k$ and $$|Q| \le (1+\ln|\mathbb{D}|)r^*+d$$ where $|\mathbb{D}|\le(\gamma+1)^{d-1}+m$ and $r^*$ is the minimum size of sets with a rank-regret at most $k$ for $\mathbb{S}$. 
\end{theorem}
\begin{IEEEproof}
Let $S^*$ be the minimum subset of $D$ with a rank-regret at most $k$ for $\mathbb{S}$, i.e., $$S^* = \arg\min_{S\subseteq D: \nabla_\mathbb{S}(S)\le k} |S|. $$ The size of $S^*$ is $r^*$. Correspondingly, $Q^*$ is the minimum set with a rank-regret at most $k$ for $\mathbb{D}_k$. Since $\mathbb{D}_k$ is a subset of $\mathbb{S}$, we have $\nabla_{\mathbb{D}_k}(S^*)\le\nabla_\mathbb{S}(S^*)\le k$, and then $|Q^*|\le|S^*|$. $Q\backslash B$ is a approximate solution for $\mathbb{D}_k$ with a $(1+\ln|\mathbb{D}_k|)$ approximate ratio on size, and $$|Q\backslash B|\le(1+\ln|\mathbb{D}_k|)|Q^*|\le(1+\ln|\mathbb{D}_k|)r^*. $$ Further, the size of $B$ is $d$ and $\mathbb{D}_k\subseteq\mathbb{D}$. In short, $|Q|\le(1+\ln|\mathbb{D}|)r^*+d$.
\end{IEEEproof}

\begin{theorem}
	\label{t_RRM_quality}
	Given a size bound $r\ge1$, a sample size $m=O(\frac{r\ln n}{\delta^2})$ and a discretization parameter $\gamma = O(\frac{1}{\epsilon})$, HDRRM returns a set $R$ such that 
	\begin{itemize}
		\item [(1)]
		$\nabla_\mathbb{D}(R)\le k^*$ and $|R|\le r$;
		\item [(2)]
		for any user, with probability at least $1-\delta$, $R$ has a rank-regret at most $k^*$ w.r.t. his/her utility function;
		\item [(3)] 
		$\forall u\in\mathbb{S}$, $w(u, R)\ge (1-\epsilon)w_{k^*}(u,D)$,
	\end{itemize}
	where $|\mathbb{D}|\le(\gamma+1)^{d-1}+m$ and $k^*$ is the minimum rank-regret for $\mathbb{S}$ of size $\frac{r-d}{1+\ln|\mathbb{D}|}$ sets.
\end{theorem}
\begin{IEEEproof}
	Based on the definition of $k^*$, there is a set $S\subseteq D$ such that $|S|=\frac{r-d}{1+\ln|\mathbb{D}|}$ and $\nabla_\mathbb{S}(S) = k^*$. And then the minimum size of sets with rank-regrets at most $k^*$ for $\mathbb{S}$ is no more than $\frac{r-d}{1+\ln|\mathbb{D}|}$. Based on Theorem \ref{t_RRR_quality} and its proof, we know that if input parameter $k=k^*$, then ASMS returns a set $Q\subseteq D$ such that $|Q|\le r$. Further, $R$ is the result of the call to ASMS, in which parameter $k$ is equal to the smallest value $k'$ in $[n]$ that makes ASMS output a set no greater than $r$. There is a trade-off between the input parameter $k$ and the size of output in ASMS. Therefore, we have $$\nabla_{\mathbb{D}}(R)\le k'\le k^*. $$ And condition (1) holds. Based on Theorem \ref{t_guarantee_1}, since $B\subseteq R$, condition (3) holds. 
	
	Let $\mathcal{K}$ be the set $[n-r+1]$. Let $\mathcal{S}$ be the collection of supersets of $B$ no larger than $r$. $R$ has a rank-regret at most $k^*\in\mathcal{K}$ for $\mathbb{D}$ and $R\subseteq\mathcal{S}$. Set $m$ to $\frac{(r-d)\ln(n-d)+\ln(n-r+1)+\ln n}{2(\delta-\frac{1}{n})^2}$. Based on Theorem \ref{t_guarantee_2}, with probability at least $1-\frac{1}{n}$, $\mathrm{Rat}_k(Q)\ge 1-\delta+\frac{1}{n}$. Under the previous assumption, each utility vector in $\mathbb{S}$ is equally probable to be used by a user. Totally, for any user, with a probability at least $(1-\delta+\frac{1}{n})(1-\frac{1}{n})>1-\delta$, there is a tuple $t$ in $Q$ ranked in the top-$k$ of $T$. Condition (2) holds. 
\end{IEEEproof}
\begin{theorem}
	\label{t_RRM_time}
	HDRRM is an $O(n(\gamma^{d-1}+m)(\log n+d))$ time and $O(n(\gamma^{d-1}+m))$ space algorithm.
\end{theorem}
\begin{IEEEproof}
	$k^*\le n$ is the minimum rank-regret for $\mathbb{S}$ of all size $\frac{r-d}{1+\ln|\mathbb{D}|}$ sets. It takes $O(n)$ and $O(|\mathbb{D}|)$ time to calculate $B$ and $\mathbb{D}$, respectively. For each $u\in\mathbb{D}$, consider precomputing the utilities of tuples w.r.t. $u$ and sorting them. The above process totally takes $O(|\mathbb{D}|n(d+\log n))$ time. Then each call of ASMS only takes $O(|\mathbb{D}|n)$ time (line 3 takes $O(k)$ time). HDRRM calls ASMS $O(\log k^*)$ times. Thus HDRRM takes $O(|\mathbb{D}|n(d+\log n+\log k^*))$ time in total. 
	The space overhead mainly comes from the precomputed total of $|\mathbb{D}|$ ordered utility lists and at most $n$ sets in $\mathcal{V}$. Each list requires $O(n)$ space, and each set requires at most $O(\mathbb{D})$ space.
\end{IEEEproof}

Note that the size bound $r$ is a constant in practice. Assuming both $\frac{1}{\epsilon}$ and $\frac{1}{\delta}$ in Theorem \ref{t_RRM_quality} are constants, then HDRRM takes $O(n\log^2 n)$ time and $O(n\log n)$ space. And it approximates the best rank-regret for $O(\frac{r}{\ln\ln n})$ size sets. 

\subsubsection{Comparison}
\label{s_hd_com}
Asudeh et al. \cite{asudeh2019rrr} proposed three approximation algorithms for RRR: MDRRR, an $O(|W|kn\mathrm{LP}(d, n))$ time and $O(|W|k)$ space (the lower-bound of $|W|$ is $n^{d-1}e^{\Omega(\sqrt{\log n})}$\cite{toth2000point}) algorithm that guarantees a logarithmic approximation-ratio on size and a rank-regret of $k$; MDRRR$_r$, the randomized version of MDRRR with $O(|W|(nd+k\log k))$ time and $O(|W|k)$ space; MDRC, a heuristic algorithm based on function space partitioning. Combined with a binary search, the above algorithms can be applied for RRM. However, compared with HDRRM, the algorithms have some serious flaws. MDRRR is a theoretical algorithm and does not scale beyond a few hundred tuples. MDRRR$_r$ runs faster but still does not handle more than $10^5$ tuples or $5$ attributes. As a sacrifice for speed, both MDRRR$_r$ and MDRC have no guarantee on rank-regret. In addition, MDRRR and MDRC are not applicable for RRRM. HDRRM does not have these defects and always has the minimal rank-regret in experiments. In contrast, MDRC often has the maximum one.  In some cases, the output rank-regret of MDRC is more than 2 orders of magnitude worse than HDRRM, which is unacceptable. Table \ref{tab_compare} summarizes the above comparisons.

\begin{table}[t]
	\caption{Comparison for Algorithms in HD}
	\centering
	\footnotesize
	\label{tab_compare}
	\begin{tabular}{p{29mm}<{\centering}|p{9mm}<{\centering}|p{10mm}<{\centering}|p{7mm}<{\centering}|p{9mm}<{\centering}} \hline
		\diagbox[width=33mm]{Criteria}{Algorithms} & MDRRR & MDRRR$_r$ & MDRC & HDRRM \\ \hline
		Guarantee on rank-regret & Yes & No & No & Yes \\ \hline
		Suitable for RRRM  & No & Yes & No & Yes \\ \hline
		Scalable for large $n$, $d$ & No & No & Yes & Yes \\ \hline
		Acceptable rank-regret & Yes & Yes & No & Yes \\ \hline
	\end{tabular}
	\vspace{-3ex} 
\end{table}

\subsection{Modifications for Other Distribution or RRRM}
Similar to \cite{xie2020being}, although it is assumed that vectors in $\mathbb{S}$ have the same probability of being used by a user, HDRRM can generalize to any other distribution through some modifications. First, the samples in $\mathbb{D}_a$ are generated based on the specific distribution of $\mathbb{S}$ instead of a uniform distribution. Second, $\mathrm{Rat}_k(\cdot)$ in Theorem \ref{t_guarantee_2} is defined by integral on $\mathbb{S}$ w.r.t. probability density, rather than the surface area. 

To generalize to RRRM, HDRRM requires the following modifications. Firstly, the $m$ vectors in $\mathbb{D}_a$ are sampled from $\mathbb{U}$. Correspondingly, $\mathrm{Rat}_k(\cdot)$ is defined by integral over $\mathbb{U}$. Secondly, the algorithm has to discard some of vectors in $\mathbb{D}_b$. Specifically, for each $u$ in $\mathbb{D}_b$, it keeps $u$ in $\mathbb{D}$, if there is a vector in $\mathbb{U}$ with the same direction as $u$, i.e., $\exists c>0$, $cu\in\mathbb{U}$. Here, the algorithm no longer calculates a normalized space of $\mathbb{U}$, like $\mathbb{P}$ in Section \ref{s_2d}.

%% file: file/experiments.tex
\section{Experiments}

All algorithms were implemented in C++ and run on a Core-I7 machine running Ubuntu 18.04 with 128 GB of RAM. Most experimental settings follow those in \cite{asudeh2017efficient,asudeh2019rrr}. Some results are plotted in log-scale for better visualization. 

There are six datasets used in experiments, three synthetic datasets (independent, correlated and anti-correlated) and three real datasets (Island, NBA, and Weather). We generate the synthetic datasets by a generator proposed by Borzsony et. al. \cite{borzsony2001skyline}. In general, the more correlated the attributes, the smaller the output rank-regrets. The real data includes three publicly available datasets commonly used in the existing researches, the Island dataset \cite{xie2020being, wang2021interactive}, the NBA dataset\cite{wang2021fully} and the Weather dataset \cite{wang2021interactive}. Island contains 63,383 2-dimensional tuples, which characterize geographic positions. NBA is a basketball dataset with 21,961 tuples, each of which represents one player/season combination on 5 attributes. Weather includes 178,080 tuples described by 4 attributes. For all datasets, the range of each attribute is normalized to $[0, 1]$. 

In 2D space, we evaluated the performance of our algorithm 2DRRM and compared it with 2DRRR proposed in \cite{asudeh2019rrr}. 2DRRM returns the optimal solution for RRM. And 2DRRR is an approximation algorithm for RRR\cite{asudeh2019rrr}. Given a threshold $k$, 2DRRR relaxes the rank-regret bound to $2k$ for the optimal output size. 
It is adapted to RRM by performing a binary search on $k$ in $[n]$ to find the smallest value that guarantees the size bound $r$. Due to the low output rank-regrets, specifically, the improved binary search in Section \ref{s_hd} is adopted to reduce the number of rounds. For lack of space, the evaluations are focused on the running time. Our algorithm 2DRRM always guarantees optimality on rank-regret. 

In HD space, we compared the performance of our algorithm HDRRM with the HD algorithms proposed in \cite{asudeh2019rrr} for RRR, namely MDRRR, MDRRR$_r$ and MDRC. Similar to 2DRRR, they are applied to RRM with a binary search. Due to the sensitivity to $k$, MDRRR and MDRRR$_r$ are enhanced by the improved binary search. The results of MDRRR are not included here, since it is a theoretical algorithm and does not scale beyond a few hundred tuples. As shown in Section \ref{s_def}, the existing algorithms proposed for RMS\cite{nanongkai2010regret} pursue the optimization on the regret-ratio, possibly resulting in a significant increase on rank-regret. To verify this, following the setting in \cite{asudeh2019rrr}, compare HDRRM with the MDRMS algorithm proposed in \cite{asudeh2017efficient}. Both the rank-regret and execution time of an algorithm are used to measure its performance. Computing the exact rank-regret of a set is not scalable to the large settings. Similar to \cite{asudeh2019rrr}, draw 100,000 functions uniformly at random and consider them for estimating the rank-regret. 

\subsection{2D Experimental Result}
\begin{figure}[tbp]
	\centering
	\begin{minipage}[t]{0.47\linewidth}
		\centering
		\includegraphics[width=1\linewidth]{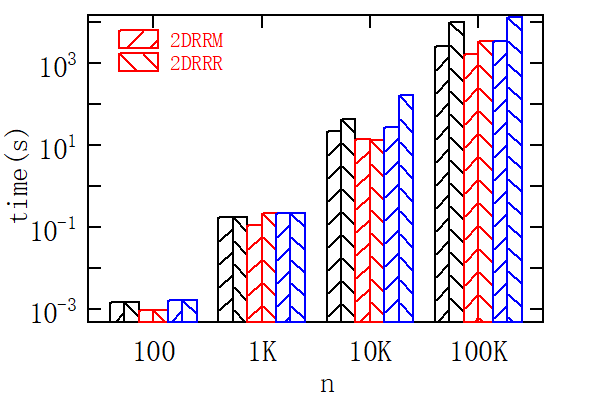} 
		\caption{2D, impact of dataset size on three synthetic datasets}
		\label{ex-2d-n-t}
	\end{minipage}
	\hspace{0.02\linewidth}
	\begin{minipage}[t]{0.47\linewidth}
		\centering
		\includegraphics[width=1\linewidth]{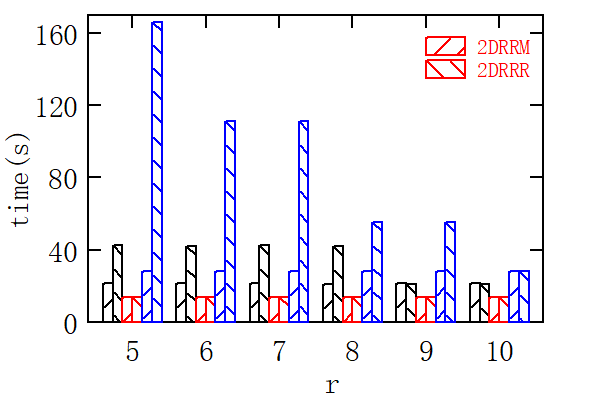} 
		\caption{2D, impact of output size on three synthetic datasets}
		\label{ex-2d-r-t}
	\end{minipage}
	\begin{minipage}[t]{0.47\linewidth}
		\centering
		\includegraphics[width=1\linewidth]{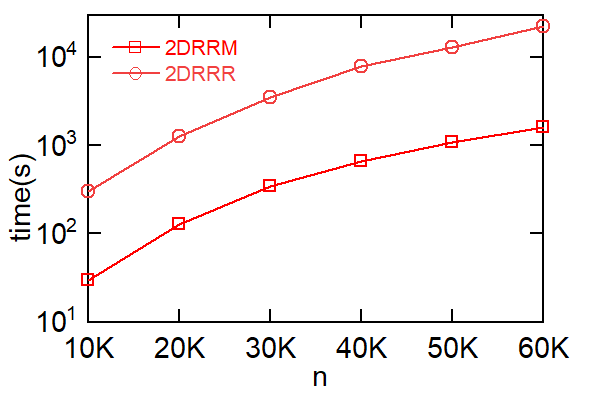} 
		\caption{2D, varied the dataset size on Island}
		\label{ex-2d-island-n-t}
	\end{minipage}
	\hspace{0.02\linewidth}
	\begin{minipage}[t]{0.47\linewidth}
		\centering
		\includegraphics[width=1\linewidth]{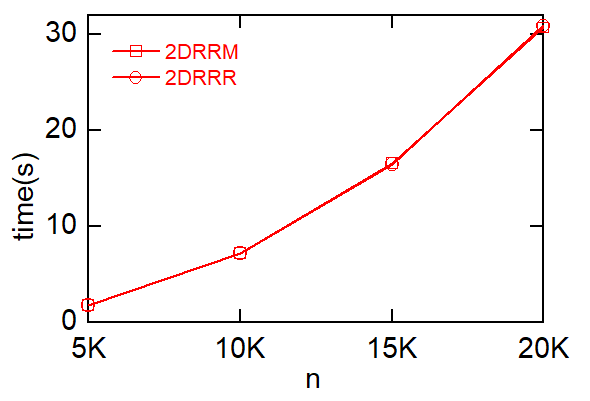} 
		\caption{2D, varied the dataset size on NBA}
		\label{ex-2d-nba-n-t}
	\end{minipage}
	\vspace{-3ex} 
\end{figure}

Figures \ref{ex-2d-n-t}, \ref{ex-2d-r-t}, \ref{ex-2d-island-n-t} and \ref{ex-2d-nba-n-t} show the performance of our algorithm 2DRRM and the algorithm 2DRRR proposed in \cite{asudeh2019rrr}. In Figures \ref{ex-2d-n-t} and \ref{ex-2d-r-t}, black, red, and blue bars are used for the independent, correlated and anti-correlated datasets respectively. The default values for the dataset size and output size are set to $n =$ 10K (K $=10^3$) and $r = 5$. Note that 2DRRM is an exact algorithm and 2DRRR is an approximae algorithm.

\subsubsection{Impact of the dataset size ($n$)}
We tested each algorithm over the independent, correlated and anti-correlated datasets. We varied the dataset size from 100 to 100K. Figure \ref{ex-2d-n-t} shows the execution time of both algorithms. As the amount of data increases, it becomes more and more obvious that 2DRRM runs faster than 2DRRR. This is because increasing the dataset size leads to an increase in the output rank-regret, and then 2DRRR requires more rounds of binary search. Compared to 2DRRR, 2DRRM is less sensitive to the correlations of attributes, since it mainly considers all intersections between lines in dual space. 2DRRR performs much worse for the anti-correlated dataset, because anti-correlations cause a higher rank-regret of output. In addition, 2DRRM outperforms 2DRRR by 4 times on the anti-correlated dataset with $n=$ 100K.

\subsubsection{Impact of the output size ($r$)}
We varied the output size from 5 to 10. Figure \ref{ex-2d-r-t} shows that as the output size increases, 2DRRM is basically unaffected but 2DRRR runs faster than before. The time complexity of 2DRRM has nothing to do with $r$. The increase in $r$ leads to a decrease in the output rank-regret, and then 2DRRR requires less rounds. Further, 2DRRM is 5 times faster than 2DRRR on the anti-correlated dataset with $r=5$.

\subsubsection{Real datasets}
Figures \ref{ex-2d-island-n-t} and \ref{ex-2d-nba-n-t} show the execution time of 2DRRM and 2DRRR over two real datasets, Island and NBA, respectively. We varied the dataset size from 10K to 60K for Island, and from 5K to 20K for NBA. Figure \ref{ex-2d-island-n-t} shows that 2DRRM outperforms 2DRRR by one order of magnitude on Island. While in Figure \ref{ex-2d-nba-n-t} the two algorithm are almost indistinguishable. This is because the output rank-regrets remain 1 on NBA. And the two algorithms have similar structure and running time. 

\subsection{HD Experimental Result}
\label{s_ex_hd}

\begin{figure*}[htbp]
	\centering
	\begin{minipage}[t]{0.23\linewidth}
		\centering
		\includegraphics[width=1\linewidth]{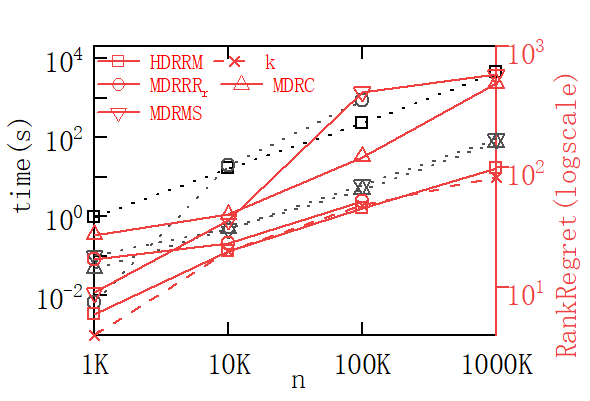} 
		\caption{HD, impact of dataset size on independent dataset}
		\label{ex-indep-n-tr}
	\end{minipage}
	\hspace{0.01\linewidth}
	\begin{minipage}[t]{0.23\linewidth}
		\centering
		\includegraphics[width=1\linewidth]{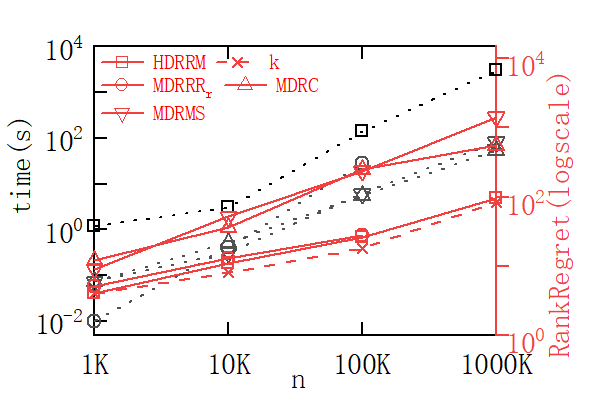} 
		\caption{HD, impact of dataset size on correlated dataset}
		\label{ex-corr-n-tr}
	\end{minipage}
	\hspace{0.01\linewidth}
	\begin{minipage}[t]{0.23\linewidth}
		\centering
		\includegraphics[width=1\linewidth]{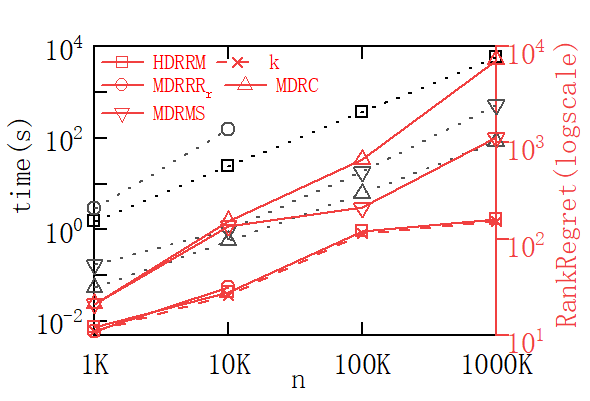} 
		\caption{HD, impact of dataset size on anti-correlated dataset}
		\label{ex-anti-n-tr}
	\end{minipage}
	\hspace{0.01\linewidth}
	\begin{minipage}[t]{0.23\linewidth}
		\centering
		\includegraphics[width=1\linewidth]{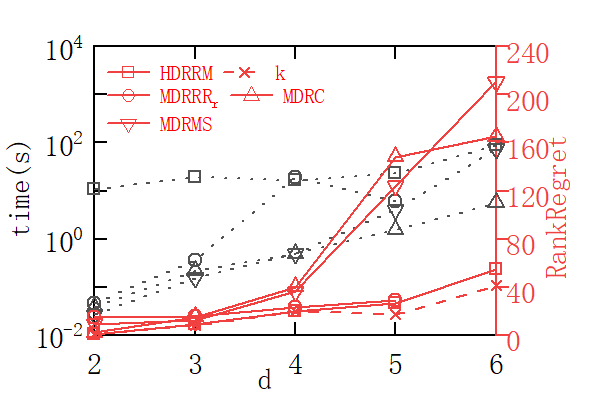} 
		\caption{HD, impact of dimension on independent dataset}
		\label{ex-indep-d-tr}
	\end{minipage}
	\begin{minipage}[t]{0.23\linewidth}
		\centering
		\includegraphics[width=1\linewidth]{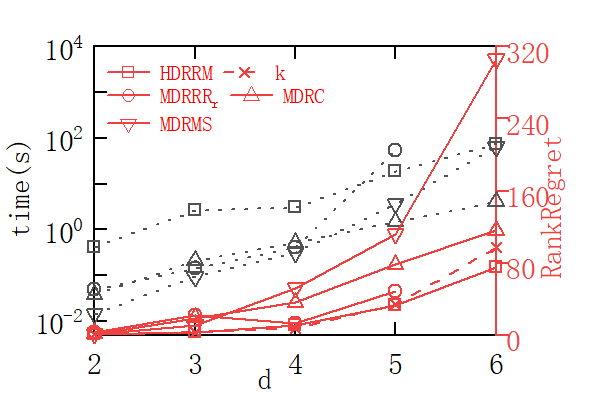} 
		\caption{HD, impact of dimension on correlated dataset}
		\label{ex-corr-d-tr}
	\end{minipage}
	\hspace{0.01\linewidth}
	\begin{minipage}[t]{0.23\linewidth}
		\centering
		\includegraphics[width=1\linewidth]{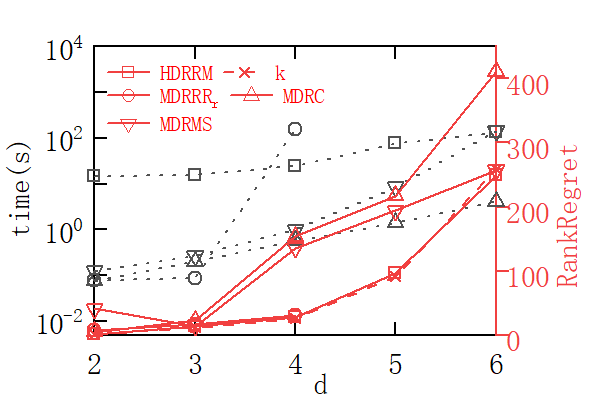} 
		\caption{HD, impact of dimension on anti-correlated dataset}
		\label{ex-anti-d-tr}
	\end{minipage}
	\hspace{0.01\linewidth}
	\begin{minipage}[t]{0.23\linewidth}
		\centering
		\includegraphics[width=1\linewidth]{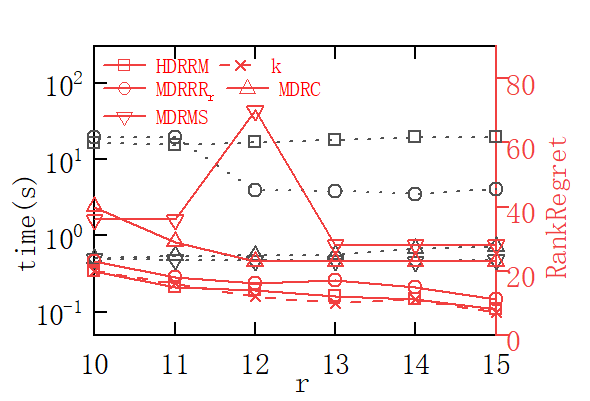} 
		\caption{HD, impact of output size on independent dataset}
		\label{ex-indep-r-tr}
	\end{minipage}
	\hspace{0.01\linewidth}
	\begin{minipage}[t]{0.23\linewidth}
		\centering
		\includegraphics[width=1\linewidth]{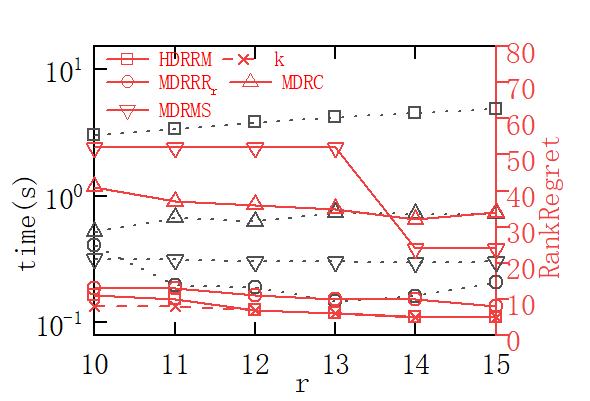} 
		\caption{HD, impact of output size on correlated dataset}
		\label{ex-corr-r-tr}
	\end{minipage}
\vspace{-3ex} 
\end{figure*}
The performance of HD algorithms was studied on the three synthetic datasets (independent, correlated and anti-correlated) and two real datasets (NBA and Weather). The default values for the dataset size, number of attributes and output size are set to $n =$ 10K, $d = 4$ and $r = 10$, respectively. Based on the proof of Theorem \ref{t_RRM_quality} (shown in \cite{xiao2021rankregret}), the sample size is set to 
\vspace{-1ex}$$m = \frac{(r-d)\ln(n-d)+\ln(n-r+1)+\ln n}{2(\delta-\frac{1}{n})^2}, \vspace{-1ex}$$
where we set $\delta=0.03$ by default. Following the setting in \cite{asudeh2017efficient}, the discretization parameter $\gamma$ is set to 6. In Figures \ref{ex-indep-n-tr} to \ref{ex-hd-weather-n-tr}, the rectangle, circle, triangle, and inverted triangle markers represent algorithms HDRRM, MDRRR$_r$, MDRC and MDRMS respectively, while the black doted lines show the execution time and the red solid lines represent the output rank-regrets. Note that the output of HDRRM is produced by a call to solver ASMS, whose input parameter $k$ (equal to the output rank-regret of HDRRM for $\mathbb{D}$) is denoted by a red cross dashed line in each figure. The red rectangle solid line represents the rank-regret of HDRRM for $\mathbb{L}$. Figures \ref{ex-indep-n-tr} to \ref{ex-hd-weather-n-tr} show that the two lines basically fit, indicating that the finite vector set $\mathbb{D}$ approximates the infinite vector space $\mathbb{L}$ well. 

Combined with a binary search for solving RRM, both MDRRR$_r$ and MDRC have no theoretical guarantee on rank-regret. In addition, MDRRR$_r$ is not scalable beyond 100K tuples or 5 attributes, as we shall show. And MDRC cannot be applied to the RRRM problem. The output of HDRRM always has the minimal rank-regret. Instead, for all experiments, either MDRMS or MDRC has the worst output quality. And in some cases, MDRMS and MDRC have rank-regrets more than one and two orders of magnitude larger than HDRRM respectively. Since a different optimization objective (i.e., the regret-ratio), MDRMS fails to have a reasonable output rank-regret.

\subsubsection{Impact of the dataset size ($n$)}
We varied the dataset size from 1K to 1000K and evaluated the performance of HDRRM, MDRRR$_r$, MDRC and MDRMS. Figures \ref{ex-indep-n-tr}, \ref{ex-corr-n-tr} and \ref{ex-anti-n-tr} show the results for the independent, correlated and anti-correlated datasets. As the amount of data increases, the running time and output rank-regrets of the four algorithms increase significantly. MDRRR$_r$ does not scale beyond 10K tuples for the anti-correlated dataset, and 100K tuples for the independent and correlated datasets. The reason is that MDRRR$_r$ needs the collection of $k$-sets, whose size grows super linearly as the data volume increases. Correspondingly, the time complexity of HDRRM is nearly linearly related to the amount of data, thus it is scalable for large $n$. On all three datasets, MDRC and MDRMS perform well in terms of running time. However, their outstanding performance is at the expense of result quality. For all settings, either MDRC or MDRMS has the highest output rank-regret. Relatively, our algorithm HDRRM always has the lowest one. On the anti-correlated dataset with $n=$ 1000K, the rank-regret of HDRRM is 160, but that of MDRC is an astonishing 7231, which is unacceptable. MDRC cannot guarantee the output quality. For the correlated dataset and the same $n$, the rank-regret of MDRMS is 1439 (that of HDRRM is 94). MDRMS outputs a set with a high rank-regret.

\subsubsection{Impact of the number of attributes ($d$)}
We varied the number of attributes from 2 to 6. Figures \ref{ex-indep-d-tr}, \ref{ex-corr-d-tr} and \ref{ex-anti-d-tr} show the results for the independent, correlated and anti-correlated datasets. As the number of attributes increases, the execution time and output rank-regret of each algorithm increase. However, compared to the others, the running time of MDRRR$_r$ increases more drastically, and it does not scale beyond 5 attributes for the independent and correlated datasets. Further, MDRRR$_r$ is sensitive to anti-correlations between attributes and cannot handle more than 4 attributes for the anti-correlated dataset. As expected, for all three datasets, HDRRM always has the minimal rank-regret. On the contrary, either MDRC or MDRMS has the worst output quality. 

\subsubsection{Impact of the output size ($r$)}
The output size is varied from 10 to 15. Figures \ref{ex-indep-r-tr}, \ref{ex-corr-r-tr} and \ref{ex-anti-r-tr} show the results for the independent, correlated and anti-correlated datasets. The execution time of HDRRM, MDRC and MDRMS is basically unaffected by the output size. But as the output size increases, MDRRR runs significantly faster. This is due to the fact that an increase in the output size leads to a decrease in rank-regret, thereby reducing the number of $k$-sets. However, not for all algorithms, the rank-regret decreases as output size increases. The rank-regret of MDRMS fluctuates up and down in the independent and anti-correlated datasets. This is because MDRMS has a different optimization objective, the regret-ratio. HDRRM always has the lowest rank-regret. And either MDRC or MDRMS has the highest one.

\subsubsection{Impact of the sample size}
\begin{figure*}[htbp]
	\centering
	\begin{minipage}[t]{0.23\linewidth}
		\centering
		\includegraphics[width=1\linewidth]{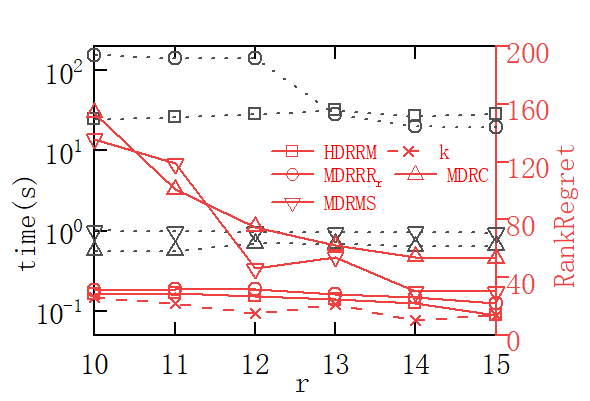} 
		\caption{HD, impact of output size on anti-correlated dataset}
		\label{ex-anti-r-tr}
	\end{minipage}
	\hspace{0.01\linewidth}
	\begin{minipage}[t]{0.23\linewidth}
		\centering
		\includegraphics[width=1\linewidth]{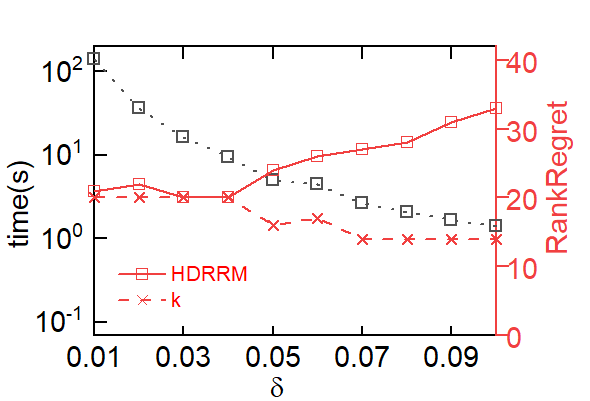} 
		\caption{HD, impact of $\delta$ on independent dataset}
		\label{ex-indep-m-tr}
	\end{minipage}
	\hspace{0.01\linewidth}
	\begin{minipage}[t]{0.23\linewidth}
		\centering
		\includegraphics[width=1\linewidth]{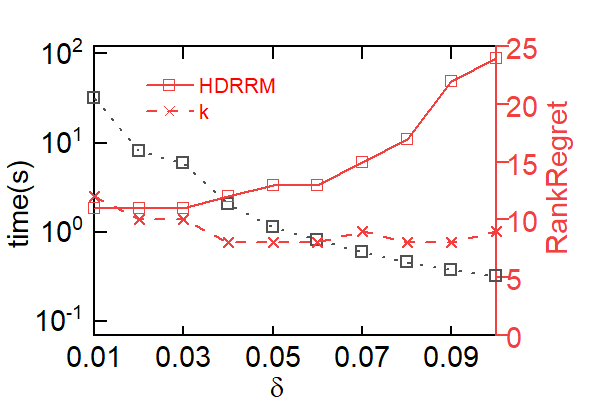} 
		\caption{HD, impact of $\delta$ on correlated dataset}
		\label{ex-corr-m-tr}
	\end{minipage}
	\hspace{0.01\linewidth}
	\begin{minipage}[t]{0.23\linewidth}
		\centering
		\includegraphics[width=1\linewidth]{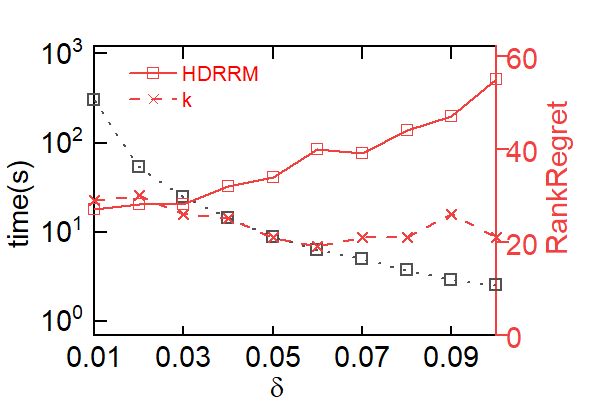} 
		\caption{HD, impact of $\delta$ on anti-correlated dataset}
		\label{ex-anti-m-tr}
	\end{minipage}
	\begin{minipage}[t]{0.23\linewidth}
		\centering
		\includegraphics[width=1\linewidth]{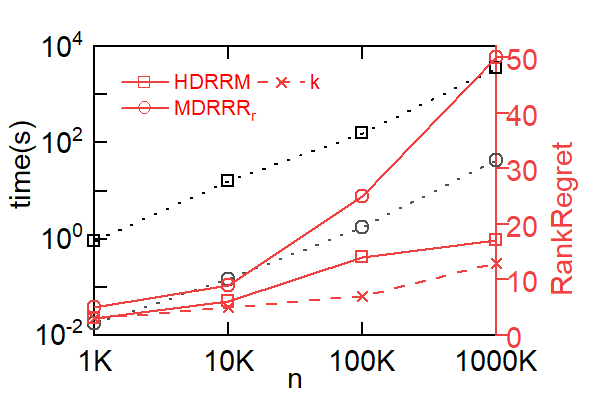} 
		\caption{HD, RRRM, varied dataset size on anti-correlated dataset}
		\label{ex-hd-rrrm-anti-n-tr}
	\end{minipage}
	\hspace{0.01\linewidth}
	\begin{minipage}[t]{0.23\linewidth}
		\centering
		\includegraphics[width=1\linewidth]{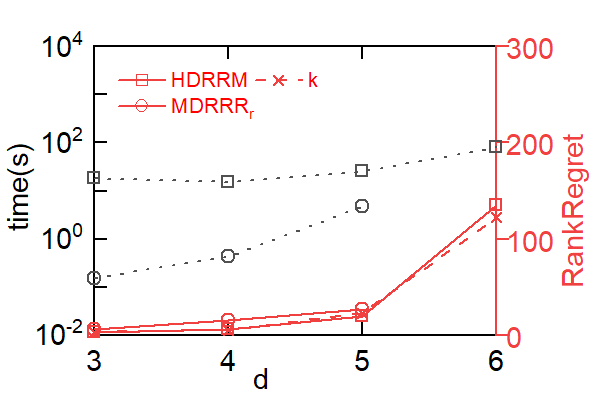} 
		\caption{HD, RRRM, varied dimension on anti-correlated dataset}
		\label{ex-hd-rrrm-anti-d-tr}
	\end{minipage}
	\hspace{0.01\linewidth}
	\begin{minipage}[t]{0.23\linewidth}
		\centering
		\includegraphics[width=1\linewidth]{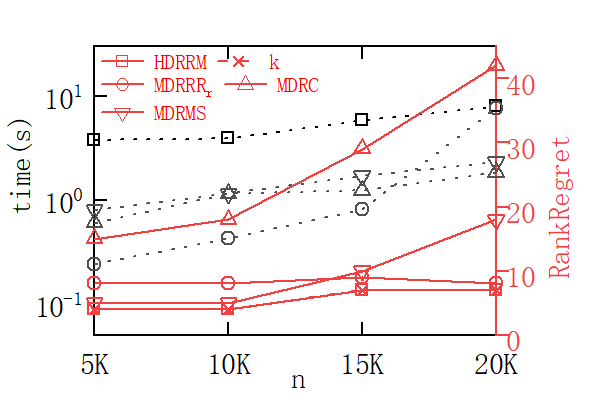} 
		\caption{HD, varied dataset size on NBA}
		\label{ex-hd-nba-n-tr}
	\end{minipage}
	\hspace{0.01\linewidth}
	\begin{minipage}[t]{0.23\linewidth}
		\centering
		\includegraphics[width=1\linewidth]{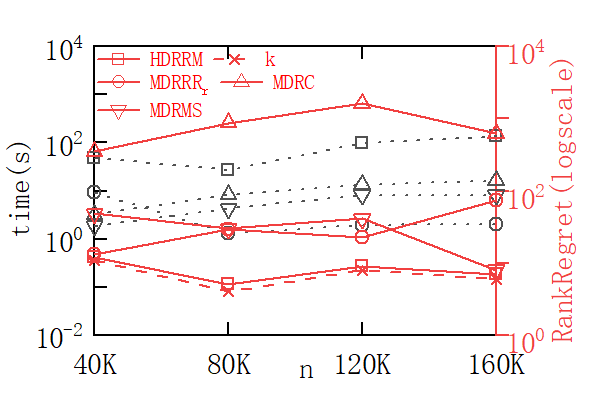} 
		\caption{HD, varied dataset size on Weather}
		\label{ex-hd-weather-n-tr}
	\end{minipage}
	\vspace{-3ex} 
\end{figure*}

Since the sample size $m$ is closely related to the error parameter $\delta$, we varied $\delta$ from $0.01$ to $0.1$. We studied the effect of the value of $\delta$ on the performance of HDRRM. Figures \ref{ex-indep-m-tr}, \ref{ex-corr-m-tr} and \ref{ex-anti-m-tr} show the results for the independent, correlated and anti-correlated datasets respectively. Increasing $\delta$ decreases the sample size and causes a rapid decrease in running time. On the other hand, it leads to an increase of the output rank-regret for HDRRM. Looking at the figures, setting $\delta=0.03$ seems appropriate, as it seems to reach a point of saturation where decreasing $\delta$ benefits little in terms of rank-regret but causes a significant increase in running time.

\subsubsection{Generalization to the RRRM problem}
We follow the setting in \cite{ciaccia2017reconciling} and let the restricted vector space be 
\vspace{-1ex}$$\mathbb{U} = \{u\in\mathbb{R}^d_+ \ |\ \forall i\in[c], u[i]\ge u[i+1]\},\vspace{-1ex}$$
which represents one of the most common types of constraints on utility vectors: weak rankings \cite{eum2001establishing}. And we set $c$ to 2. For lack of space, we only show the results on the anti-correlated dataset, which is the most interesting synthetic dataset\cite{xie2020experimental}. In Figure \ref{ex-hd-rrrm-anti-n-tr}, we varied the dataset size from 1K to 1000K and evaluated the performance of HDRRM and MDRRR$_r$. Since the vector space is severely restricted, the number of $k$-sets decreases and MDRRR$_r$ runs faster than before. However, as the increase of dataset size, the gap between the rank-regret of MDRRR$_r$ and that of HDRRM becomes more obvious. In Figure \ref{ex-hd-rrrm-anti-d-tr}, we varied the number of attributes from 3 to 6. As the dimension increases, the execution time and output rank-regret of each algorithm increase. However, compared with HDRRM, MDRRR$_r$ is more sensitive to the dimension and does not scale beyond 5 attributes. Further, HDRRM always has the minimal rank-regret. 

\subsubsection{Real datasets}
We also evaluated the performance of HD algorithms over the NBA and Weather datasets. In Figure \ref{ex-hd-nba-n-tr}, we varied the dataset size from 5K to 20K and set the number of attributes to 5 for NBA. MDRC and MDRMS perform well in terms of running time. And the execution time of MDRRR$_r$ is most affected by the amount of data. When $n=$ 20K, HDRRM and MDRRR$_r$ are equally efficient. HDRRM and MDRC have the lowest and highest rank-regrets, respectively. In Figure \ref{ex-hd-weather-n-tr}, we varied the dataset size from 40K to 160K and set the dimension to 4 for Weather. The results are basically the same as before, but the output quality of MDRC is much more terrible. When $n=$ 120K, the output rank-regret of HDRRM is 9, but that of MDRC is an unbelievable 1610. There is a difference of more than two orders of magnitude between them. Even for other data volumes, the difference is more than one order. MDRC is not adapted to the Weather dataset.

\subsection{Summary of results}
To summarize, the experiments verified the effectiveness and efficiency of our algorithms 2DRRM and HDRRM. 2DRRM returns the optimal solution and runs faster than the approximate algorithm 2DRRR. Our HD algorithm HDRRM also outperforms the other algorithms. HDRRM always has the lowest output rank-regret. Further, HDRRM is applicable for the RRRM problem and is scalable for large data volumes and high dimensions. While MDRC runs faster, its output quality is often the worst and in some cases, is even unacceptable, which is more than 2 orders of magnitude worse than HDRRM. In addition, MDRC is not suitable for the RRRM problem. Although MDRRR$_r$ can be applied to RRRM, it does not scale beyond 100K tuples or 5 attributes. Further, both MDRRR$_r$ and MDRC have no guarantee on rank-regret. MDRMS performs well in terms of execution time, but it pursues the optimization on the regret-ratio\cite{nanongkai2010regret}, possibly resulting in a significant increase on rank-regret. Sometimes MDRMS fails to have a reasonable output rank-regret. Experiments verified that it is feasible to approximate the full vector space $\mathbb{L}$ with the discrete vector set $\mathbb{D}$.

%% file: file/conclusion.tex
\section{Conclusion}
In this paper, we generalize RRM and propose the RRRM problem to find a set, limited to only $r$ tuples, that minimizes the rank-regret for utility functions in a restricted space. The solution of RRRM usually has a lower regret level and can better serve the specific preferences of some users. We make several theoretical results as well as practical advances for RRM and RRRM. We prove that both RRM and RRRM are shift invariant. With a theoretical lower-bound, we show that there is no algorithm with a upper-bound, independent of the data volume, on the rank-regret. In 2D space, we design the algorithm 2DRRM to find the optimal solution for RRM. 2DRRM can be applied to the RRRM problem. In HD space, we propose the algorithm HDRRM, the only HD algorithm that has a theoretical guarantee on rank-regret and is applicable for RRRM at the same time. HDRRM always has the best output quality in experiments. The comprehensive set of experiments on the three synthetic datasets and the three real datasets verify the efficiency, output quality, and scalability of our algorithms.

%% file: main.bbl
\begin{thebibliography}{10}

\bibitem{agarwal2017efficient}
Pankaj~K. Agarwal, Nirman Kumar, Stavros Sintos, and Subhash Suri.
\newblock Efficient algorithms for k-regret minimizing sets.
\newblock In {\em 16th International Symposium on Experimental Algorithms, June
  21-23, 2017, London, {UK}}, volume~75 of {\em LIPIcs}, pages 7:1--7:23, 2017.

\bibitem{asudeh2017efficient}
Abolfazl Asudeh, Azade Nazi, Nan Zhang, and Gautam Das.
\newblock Efficient computation of regret-ratio minimizing set: A compact
  maxima representative.
\newblock In {\em Proceedings of the 2017 ACM International Conference on
  Management of Data}, pages 821--834, 2017.

\bibitem{asudeh2019rrr}
Abolfazl Asudeh, Azade Nazi, Nan Zhang, Gautam Das, and HV~Jagadish.
\newblock Rrr: Rank-regret representative.
\newblock In {\em Proceedings of the 2019 International Conference on
  Management of Data}, pages 263--280, 2019.

\bibitem{borzsony2001skyline}
Stephan Borzsony, Donald Kossmann, and Konrad Stocker.
\newblock The skyline operator.
\newblock In {\em Proceedings 17th international conference on data
  engineering}, pages 421--430. IEEE, 2001.

\bibitem{bronnimann1995almost}
Herv{\'e} Br{\"o}nnimann and Michael~T Goodrich.
\newblock Almost optimal set covers in finite vc-dimension.
\newblock {\em Discrete \& Computational Geometry}, 14(4):463--479, 1995.

\bibitem{cao2017k}
Wei Cao, Jian Li, Haitao Wang, Kangning Wang, Ruosong Wang, Raymond
  Chi-Wing~Wong, and Wei Zhan.
\newblock k-regret minimizing set: Efficient algorithms and hardness.
\newblock In {\em 20th International Conference on Database Theory (ICDT
  2017)}. Schloss Dagstuhl-Leibniz-Zentrum fuer Informatik, 2017.

\bibitem{chester2014computing}
Sean Chester, Alex Thomo, S~Venkatesh, and Sue Whitesides.
\newblock Computing k-regret minimizing sets.
\newblock {\em Proceedings of the VLDB Endowment}, 7(5):389--400, 2014.

\bibitem{chvatal1979greedy}
Vasek Chvatal.
\newblock A greedy heuristic for the set-covering problem.
\newblock {\em Mathematics of operations research}, 4(3):233--235, 1979.

\bibitem{ciaccia2017reconciling}
Paolo Ciaccia and Davide Martinenghi.
\newblock Reconciling skyline and ranking queries.
\newblock {\em Proceedings of the VLDB Endowment}, 10(11):1454--1465, 2017.

\bibitem{de1997computational}
Mark De~Berg, Marc Van~Kreveld, Mark Overmars, and Otfried Schwarzkopf.
\newblock Computational geometry.
\newblock In {\em Computational geometry}, pages 1--17. Springer, 1997.

\bibitem{edelsbrunner2012algorithms}
Herbert Edelsbrunner.
\newblock {\em Algorithms in combinatorial geometry}, volume~10.
\newblock Springer Science \& Business Media, 2012.

\bibitem{eum2001establishing}
Yun~Seong Eum, Kyung~Sam Park, and Soung~Hie Kim.
\newblock Establishing dominance and potential optimality in multi-criteria
  analysis with imprecise weight and value.
\newblock {\em Computers \& Operations Research}, 28(5):397--409, 2001.

\bibitem{ilyas2008survey}
Ihab~F Ilyas, George Beskales, and Mohamed~A Soliman.
\newblock A survey of top-k query processing techniques in relational database
  systems.
\newblock {\em ACM Computing Surveys (CSUR)}, 40(4):1--58, 2008.

\bibitem{jamieson2011active}
Kevin~G Jamieson and Robert~D Nowak.
\newblock Active ranking using pairwise comparisons.
\newblock {\em arXiv preprint arXiv:1109.3701}, 2011.

\bibitem{joachims2002optimizing}
Thorsten Joachims.
\newblock Optimizing search engines using clickthrough data.
\newblock In {\em Proceedings of the eighth ACM SIGKDD international conference
  on Knowledge discovery and data mining}, pages 133--142, 2002.

\bibitem{liu2021eclipse}
Jinfei Liu, Li~Xiong, Qiuchen Zhang, Jian Pei, and Jun Luo.
\newblock Eclipse: Generalizing knn and skyline.
\newblock In {\em 2021 IEEE 37th International Conference on Data Engineering
  (ICDE)}, pages 972--983. IEEE, 2021.

\bibitem{mouratidis2021marrying}
Kyriakos Mouratidis, Keming Li, and Bo~Tang.
\newblock Marrying top-k with skyline queries: Relaxing the preference input
  while producing output of controllable size.
\newblock In {\em Proceedings of the 2021 International Conference on
  Management of Data}, pages 1317--1330, 2021.

\bibitem{mouratidis2018exact}
Kyriakos Mouratidis and Bo~Tang.
\newblock Exact processing of uncertain top-k queries in multi-criteria
  settings.
\newblock {\em Proceedings of the VLDB Endowment}, 11(8):866--879, 2018.

\bibitem{nanongkai2010regret}
Danupon Nanongkai, Atish~Das Sarma, Ashwin Lall, Richard~J Lipton, and Jun Xu.
\newblock Regret-minimizing representative databases.
\newblock {\em Proceedings of the VLDB Endowment}, 3(1-2):1114--1124, 2010.

\bibitem{phillips2012chernoff}
Jeff~M Phillips.
\newblock Chernoff-hoeffding inequality and applications.
\newblock {\em arXiv preprint arXiv:1209.6396}, 2012.

\bibitem{qian2015learning}
Li~Qian, Jinyang Gao, and HV~Jagadish.
\newblock Learning user preferences by adaptive pairwise comparison.
\newblock {\em Proceedings of the VLDB Endowment}, 8(11):1322--1333, 2015.

\bibitem{toth2000point}
G{\'e}za T{\'o}th.
\newblock Point sets with many k-sets.
\newblock In {\em Proceedings of the sixteenth annual symposium on
  Computational geometry}, pages 37--42, 2000.

\bibitem{wang2011latent}
Hongning Wang, Yue Lu, and ChengXiang Zhai.
\newblock Latent aspect rating analysis without aspect keyword supervision.
\newblock In {\em Proceedings of the 17th ACM SIGKDD international conference
  on Knowledge discovery and data mining}, pages 618--626, 2011.

\bibitem{wang2021interactive}
Weicheng Wang, Raymond Chi-Wing Wong, and Min Xie.
\newblock Interactive search for one of the top-k.
\newblock In {\em Proceedings of the 2021 International Conference on
  Management of Data}, pages 1920--1932, 2021.

\bibitem{wang2021fully}
Yanhao Wang, Yuchen Li, Raymond Chi-Wing Wong, and Kian-Lee Tan.
\newblock A fully dynamic algorithm for k-regret minimizing sets.
\newblock In {\em 2021 IEEE 37th International Conference on Data Engineering
  (ICDE)}, pages 1631--1642. IEEE, 2021.

\bibitem{wang2021minimum}
Yanhao Wang, Michael Mathioudakis, Yuchen Li, and Kian-Lee Tan.
\newblock Minimum coresets for maxima representation of multidimensional data.
\newblock In {\em ACM SIGACT-SIGMOD-SIGART Symposium on Principles of Database
  Systems (PODS)}, 2021.

\bibitem{xiao2020sampling}
Xingxing Xiao and Jianzhong Li.
\newblock Sampling-based approximate skyline calculation on big data.
\newblock In {\em International Conference on Combinatorial Optimization and
  Applications}, pages 32--46. Springer, 2020.

\bibitem{xiao2021rankregret}
Xingxing Xiao and Jianzhong Li.
\newblock Rank-regret minimization.
\newblock {\em CoRR}, abs/2111.08563, 2021.

\bibitem{xie2020experimental}
Min Xie, Raymond Chi-Wing Wong, and Ashwin Lall.
\newblock An experimental survey of regret minimization query and variants:
  bridging the best worlds between top-k query and skyline query.
\newblock {\em The VLDB Journal}, 29(1):147--175, 2020.

\bibitem{xie2018efficient}
Min Xie, Raymond Chi-Wing Wong, Jian Li, Cheng Long, and Ashwin Lall.
\newblock Efficient k-regret query algorithm with restriction-free bound for
  any dimensionality.
\newblock In {\em Proceedings of the 2018 international conference on
  management of data}, pages 959--974, 2018.

\bibitem{xie2020being}
Min Xie, Raymond Chi-Wing Wong, Peng Peng, and Vassilis~J Tsotras.
\newblock Being happy with the least: Achieving $\alpha$-happiness with minimum
  number of tuples.
\newblock In {\em 2020 IEEE 36th International Conference on Data Engineering
  (ICDE)}, pages 1009--1020. IEEE, 2020.

\end{thebibliography}
